\documentclass[12pt]{article}
\usepackage{amsmath, amssymb, amsthm, enumerate, graphicx, float, bm, bbm, multirow}
\usepackage[singlelinecheck = false]{caption}
\usepackage[colorlinks = true, allcolors = blue]{hyperref}
\usepackage[title]{appendix}
\usepackage[round]{natbib} 

\newcommand{\blind}{0}
\def\spacingset#1{\renewcommand{\baselinestretch}{#1}\small\normalsize}

\def\argmin{\mathop{\rm argmin}}
\def\argmax{\mathop{\rm argmax}}
\def\sign{\mathop{\rm sign}}

\newcommand{\mF}{\mathcal{F}}

\newcommand{\mX}{\mathcal{X}}
\newcommand{\mA}{\mathcal{A}}

\newcommand{\bX}{{\bm X}}
\newcommand{\bx}{{\bm x}}
\newcommand{\bZ}{{\bm Z}}
\newcommand{\bz}{{\bm z}}
\newcommand{\bbeta}{{\bm\beta}}

\newcommand{\balpha}{{\bm\alpha}}
\newcommand{\bgamma}{{\bm\gamma}}

\newcommand{\bdf}{{\bm f}}

\newcommand{\bW}{{\bm W}}
\newcommand{\be}{{\bm e}}

\newcommand{\bone}{{\bm 1}}

\newcommand{\R}{\mathbb{R}}
\newcommand{\E}{\mathbb{E}}
\newcommand{\ind}[1]{\mathbbm{1}{\left[ {#1} \right] }}

\newcommand{\<}[1]{\langle{#1}\rangle}
\newcommand{\ie}{i.e.}

\newtheorem{assumption}{Assumption}
\newtheorem{proposition}{Proposition}
\newtheorem{theorem}{Theorem}
\newtheorem{corollary}{Corollary}

\date{}

\begin{document}

\spacingset{1}
\date{}

\if0\blind
{
  \title{\bf Near-optimal Individualized Treatment Recommendations}
  \author{Haomiao Meng, Ying-Qi Zhao, Haoda Fu and Xingye Qiao\thanks{
    Correspondence to: Xingye Qiao (e-mail: qiao@math.binghamton.edu). Haomiao Meng is a PhD student in the Department of Mathematical Sciences at Binghamton University, State University of New York, Binghamton, New York, 13902;
    Yingqi Zhao is an Associate Member at Public Health Sciences Division of the Fred Hutchinson Cancer Research Center, Seattle, Washington, 98109;
    Haoda Fu is a Senior Research Advisor at Eli Lilly and Company, Indianapolis, Indiana, 46285;
    Xingye Qiao is an Associate Professor in the Department of Mathematical Sciences at Binghamton University, State University of New York, Binghamton, New York, 13902.}}
  \maketitle
} \fi

\if1\blind
{
  \bigskip
  \bigskip
  \bigskip
  \begin{center}
    {\LARGE\bf Near-optimal Individualized Treatment Recommendations}
\end{center}
  \medskip
} \fi

\bigskip
\begin{abstract}
\noindent Individualized treatment recommendation (ITR) is an important analytic framework for precision medicine. The goal is to assign proper treatments to patients based on their individual characteristics. From the machine learning perspective, the solution to an ITR problem can be formulated as a weighted classification problem to maximize the average benefit that patients receive from the recommended treatments. Several methods have been proposed for ITR in both binary and multicategory treatment setups. In practice, one may prefer a more flexible recommendation with multiple treatment options. This motivates us to develop methods to obtain a set of near-optimal individualized treatment recommendations alternative to each other, called alternative individualized treatment recommendations (A-ITR). We propose two methods to estimate the optimal A-ITR within the outcome weighted learning (OWL) framework. We show the consistency of these methods and obtain an upper bound for the risk between the theoretically optimal recommendation and the estimated one. We also conduct simulation studies, and apply our methods to a real data set for Type 2 diabetic patients with injectable antidiabetic treatments. These numerical studies have shown the usefulness of the proposed A-ITR framework. We develop a \textsf{R} package \textsf{aitr} which can be found at https://github.com/menghaomiao/aitr.
\end{abstract}

\vfill

\newpage

\spacingset{1.45} 
\setcounter{page}{1}
\abovedisplayskip=8pt
\belowdisplayskip=8pt

\section{Introduction}
The individualized treatment recommendation (ITR) has drawn increasing attentions in recent years. Because patients respond differently to the same treatment in some diseases \citep{lesko2007personalized, insel2009translating}, it is desirable to individualize the treatment according to patients' characteristics. Mathematically, an ITR is a map from the covariates to a treatment. The goal is to find the optimal ITR so that the average benefit that patients will receive by following such a recommendation is maximized.

In the literature, many statistical approaches have been proposed for solving the optimal ITR. 
For indirect modeling-based methods, one first builds a parametric or semi-parametric model to estimate the expected outcome based on a patient's characteristics, then recommends the treatment that renders the optimal outcome to the patient \citep{robins2004optimal, qian2011performance, schulte2014q}. However, they require correct model specification and accurate estimation to work well practically. One may also solve the optimal ITR directly. \cite{zhao2012estimating} proposed a classification-based method, coined as the outcome weighted learning (OWL), to estimate the optimal ITR. They transformed the ITR problem into a weighted classification problem and used support vector machine (SVM), a classification method, to solve it.
Built on top of the OWL framework, there is a rapidly growing literature on different aspects of the ITR problem. \cite{zhao2014doubly} and \cite{cui2017tree} extended the OWL framework to accommodate survival outcome. \cite{zhou2017residual} and \cite{liu2018augmented} proposed residual weighted learning (RWL) and augmented outcome-weighted learning (AOL) respectively to reduce the variability of weight in OWL to enhance its performance. \cite{chen2018estimating} proposed generalized OWL (GOWL) to solve an ITR with ordinal treatments. \cite{zhang2018multicategory} proposed angle-based approach for the multicategory case (in which there are more than two treatments to choose from). Recently \cite{zhao2019efficient} and \cite{huang2019multicategory} considered replacing the weight in OWL with a doubly-robust estimator to further improve the robustness of OWL. Methods based on other learning algorithms such as trees \citep{laber2015tree, kallus2016learning, doubleday2018algorithm, zhu2017greedy} and nearest neighbors \citep{zhou2017causal, wu2019matched} are also studied. Another example of direct-search methods is the work by \cite{zhang2012robust}, which searched for the ITR among a pre-specified class of decision rules that optimized a doubly robust augmented inverse probability weighted estimator of the overall population mean outcome.

Despite of the success of these methods in recommending a single ``optimal'' treatment to patients, a method that can suggest multiple ``near-optimal'' treatment options to a patient is not fully studied. Such options could be desirable when several treatments have comparable effects. \cite{laber2014set} and \cite{lizotte2016multi} proposed a set-valued dynamic treatment regime. In particular, if there are two treatments available ($1$ and $-1$), their set-valued rule may report $\{1\}$, $\{-1\}$, or $\{1, -1\}$. However, this approach is applicable only to cases with two competing outcomes. They would recommend the set $\{1, -1\}$ if any one treatment cannot be proven to be inferior to the other based on the two outcomes. On the other hand, they used a regression-based method to estimate the optimal set-valued rule, which may suffer an inconsistency issue if the model is mis-specified. \cite{yuan2015outcome} considered a framework to allow a reject option in ITR estimation based on OWL. However, the method is restricted to the binary case (only two possible treatments). 

In this paper, we propose to study the ITR problem in the setting with only one clinical outcome from a new perspective. Different from the previous ITR work, it provides a set of ITRs that are near the optimality and are alternative to each other, which we called alternative individualized treatment recommendations (A-ITR). Specifically, multiple treatments are recommended to the patient if they are expected to result in similar clinical outcomes for the patient. There are multiple reasons such alternative options are desired. Firstly, for some patients, since multiple treatments may yield the same or similar outcomes, the ranking between the top treatment options may vary due to some randomness or noise in the learning process. In the case that the expected outcomes for multiple outcomes are indistinguishable, it is morally inappropriate to withhold such important information from the patients. Secondly, such alternative options allow patients to incorporate other factors into their choice of the final treatment plan. These factors include the healthcare expense, the painfulness of the treatment, the life quality and life style, and so on. Specifically, when two treatments are expected to have similar outcomes, it is reasonable for the patient to choose an option which is covered by the insurance, that is less painful, or that does not significantly compromise the quality of life. In this sense, conventional ITR methods that only recommend one treatment to a patient may prevent patients from making informed decisions about their lives.

We will propose two methods to estimate A-ITR. Parallel to the development of the conventional ITR methods, we first introduce a regression-based plug-in method to estimate the optimal A-ITR, which will serve as the baseline. Within the OWL framework, we propose two classification-based methods. The technical tool we will use is multicategory classification with reject and refine options \citep{zhang2018reject}.

The rest of the paper is organized as follows. In Section \ref{sec:bg}, we review the background of the ITR and the classification with reject and refine options problems. We then introduce the proposed A-ITR framework and discuss several estimation methods in Section \ref{sec:method}. Discussions about the algorithm and the tuning procedure can be found in Section \ref{sec:opt}. In Section \ref{sec:theory}, we study the statistical learning theory for our proposed methods. Simulation studies and an application to Type 2 diabetes mellitus data are provided in Sections \ref{sec:sim} and \ref{sec:real} respectively. Some concluding remarks are given in Section \ref{sec:con}. All technical proofs are provided in the supplementary materials.

\section{Background}\label{sec:bg}
In this section we briefly review the background information of both ITR and the problem of classification with reject and refine options.
\subsection{Individualized Treatment Recommendation}\label{sec:ITR}
Denote the covariates of a patient by $\bX\in\mX$. Each treatment is denoted by a random variable $A$, where $A\in\mA=\{1,2,\dots,k\}$ ($k$ treatments available.) After assigning a treatment to a patient, we observe an outcome $Y\in\R^+$. Here we assume $Y$ is bounded and \emph{smaller} $Y$ is preferred. Then an individualized treatment recommendation, previously often referred to as an individualized treatment rule, is defined to be a map $d:\mX\to\mA$.

Let $Y^*(j)$ denote the potential outcome that would have been observed when treatment $j$ is assigned to the patient. The actual observed outcome $Y$ is related to the potential outcomes by $Y=\sum_{j\in \mA} Y^*(j) \ind{A=j}$. Define $p(A=j|\bX)$ as the conditional probability of treatment $j$ given $\bX$. We assume the following assumption.

\begin{assumption}\label{asp:reg}
For any $j$, $Y^*(j)$ is independent of $A$ given $\bX$; $p(A=j|\bX)>0$ almost everywhere.
\end{assumption}

Under Assumption \ref{asp:reg}, it was shown by \citet{qian2011performance} and \citet{kallus2016learning} that the expected outcome under ITR $d$ is
\begin{align}\label{E^d}
\E^d(Y) = \E(Y^*(d(\bX)))=\E[\E(Y^*(A)|A=d(\bX),\bX))]=\E\left[\frac{\ind{A=d(\bX)}}{p(A|\bX)}Y\right],
\end{align}
where $\E^d$ is the expectation under ITR $d$. Note that $p(A|\bX)$ is usually known in a randomized trial, while in an observational study $p(A|\bX)$ is unknown and needs to be estimated first.

Denote $\mu_j=\E(Y|\bX,A=j)$, for $j=1,\dots,k$. Then the optimal ITR $d^*$ under (\ref{E^d}) is
\begin{align}\label{opt_itr}
d^*=\argmin_d\E^d(Y)=\argmin_j\mu_j,
\end{align}
that is, the optimal treatment for a patient has the smallest (the best) expected outcome.

Many methods have been proposed for estimating the optimal ITR. One method is often called ``regression and comparison" or Q-learning \citep{robins2004optimal, qian2011performance}. One first estimates the conditional mean $\mu_j(\bx) = \E(Y \mid \bX = \bx, A = j)$ for each treatment $j$, then the optimal treatment is obtained by plugging the estimators in (\ref{opt_itr}). However, this method relies on the accuracy of the regression model. If the model is mis-specified, the error could be fairly substantial. Another group of methods treat the problem as a classification problem. One example is called outcome weight learning (OWL) or O-learning \citep{zhao2012estimating, zhao2014doubly, zhao2019efficient, zhou2017residual, zhang2018multicategory}. In the OWL framework, we rewrite the ITR solution as
\begin{align}\label{obj}
d^*=\argmin_d\E\left[\frac{Y}{p(A|\bX)}\ind{A=d(\bX)}\right],
\end{align}
which is closely related to a weighed classification problem with weight $Y/p(A|\bX)$. To overcome the non-continuity and non-convexity of the 0-1 loss, we can replace $\ind{A=d(\bX)}$ by a convex surrogate loss $L(A, \bdf(\bX))$ in the empirical counterpart and solve instead
\begin{align}\label{owl}
\hat\bdf=\argmin_\bdf\E_n\left[\frac{Y}{p(A|\bX)}L(A, \bdf(\bX))\right],
\end{align}
where $\E_n$ denotes the empirical expectation, and $\bdf$ is a multi-dimensional function defined on $\mX$. The estimated ITR $\hat d$ is then obtained from $\hat\bdf$.

The relationship between $\hat d$ and $\hat\bdf$ depends on the loss function $L$ and the choice of $\bdf$. \cite{zhao2012estimating} proposed to replace the 0-1 loss by hinge loss in the binary case ($k=2$, $A\in\{1, -1\}$), that is, $L(A, f)=(1-Af)_+$, where $x_+=\max(x, 0)$, and $f$ is a 1-dimensional function. In the current setting that a smaller $Y$ is preferred, they could have used $L(A, f)=(1+Af)_+$. They showed that the optimal ITR can be estimated by $\hat d=\sign{(\hat f)}$. \cite{zhang2018multicategory} then extended to the multicategory case using a large-margin loss and the angle-based learning framework \citep{zhang2014multicategory}. Specifically, define $\bdf(\bx)=(f_1,\dots,f_{k-1})^T(\bx)\in\R^{k-1}$ and $\bW_1, \dots,\bW_k$ are vertices of a $(k-1)$-dimensional simplex, that is
$$\bW_j=\begin{cases}
(k-1)^{-1/2}\bone_{k-1},&j=1\\
-(1+k^{1/2})(k-1)^{-3/2}\bone_{k-1}+\left[k/(k-1)\right]^{1/2}\be_{j-1},\quad&2\leq j\leq k,
\end{cases}$$
where $\bone_{k-1}$ is a $(k-1)$-dimensional vector with all $1$ and $\be_{j-1}\in\R^{k-1}$ is a vector with the $(j-1)$th element 1 and 0 elsewhere. They let $L(A, \bdf(\bx))=\ell(\<{\bW_A, \bdf(\bx)})$ where $\ell$ is a typical large-margin surrogate loss for binary classification (except that it is increasing instead of decreasing.) From the geometry point of view, treatment $j$ is represented by vertex $j$ of the simplex, and the angle between $\bdf(\bx)$ and $\bW_j$, $\angle{(\bW_j, \bdf(\bx))}$, indicates how far away $\bdf(\bx)$ is from each of these treatments. The resulting ITR was estimated by $\hat d(\bx)=\argmin_j\angle{(\bW_j, \hat\bdf(\bx))}=\argmax_j\<{\bW_j, \hat\bdf(\bx)}$, that is, the treatment whose corresponding vertex is closest to $\hat\bdf(\bx)$.

\textsc{Remark.} In the conventional ITR literature, one typically assumes that larger values of the outcome $Y$ are preferred, so that instead of minimization, $d^*$ is the solution to the maximization of the objective (\ref{obj}), or equivalently, $\argmin_d\E\left[\frac{Y}{p(A|\bX)}\ind{A\neq d(\bX)}\right],$ which was indeed a weighed classification. In this article, we assume that smaller values of $Y$ are preferred (due to a technical concern about computational complexity.) As consequences, $\ind{A\neq d(\bX)}$ is replaced by $\ind{A=d(\bX)}$ in (\ref{obj}); additionally, the surrogate loss function is flipped with respect to the origin so that it is an increasing function instead of a decreasing function.


\subsection{Classification with Reject and Refine Options}
In this article, we aim to provide set-valued recommendations that are near the optimality and are alternative to each other. We borrow the idea of multicategory classification with reject (and refine) options as a technical tool. Classification with a reject option has been widely studied. \cite{herbei2006classification} formulated the problem as a minimization problem under the 0-$d$-1 loss. That is, the loss of a misclassified instance is 1 and the loss of a rejected instance is $d$, where $0\leq d\leq1/2$. \cite{bartlett2008classification} proposed an estimation procedure under the hinge loss. \cite{yuan2010classification} extended this framework to a broad class of surrogate loss functions. \cite{zhang2018reject} generalized it to the multicategory case.

We first introduce binary classification with reject option. Let $(\bX, A)$ be a pair of random variable with $\bX\in\mX$ and class label $A\in\{1, -1\}$\footnote{Although the class label is often denoted as $Y$ in the classification literature, the role of the class label is comparable to the role of the treatment option in the ITR setting. Hence we denote the class label as $A$ here.}, and denote $p_j(\bx)=p(A=j|\bX=\bx)$ as the conditional class probability given $\bX$. The goal is to train a classifier $\phi(\bx)$ that produces three possible outputs: $1$, $-1$, and $0$. Here 0 stands for a ``reject" option, meaning that the classifier refuses to make a prediction based on the information available. Note that the decision ``0" can be viewed as a set-valued decision of $\{1,-1\}$. \cite{chow1970optimum} proposed the $0$-$d_0$-$1$ loss with corresponding risk function $P(\phi(\bX)\neq A, \phi(\bX)\neq0)+d_0P(\phi(\bX)=0)$ and it was shown that the Bayes rule under this risk is
$$\phi^*(\bx)=\begin{cases}1,&p_1(\bx)>1-d_0\\
-1,&p_1(\bx)<d_0\\
0,&d_0\leq p_1(\bx)\leq1-d_0.\end{cases}$$
Here $d_0\in[0, 1/2]$ controls the cost for refusing to make a classification. Intuitively, we produce the reject option ``0" only when both $p_1$ and $p_2$ are close to $1/2$. \cite{bartlett2008classification} proposed a bent hinge loss to estimate the optimal rule $\phi^*$. The bent hinge loss is defined as $\ell(u)=\max(0, 1-u, 1-(1-d_0)u/d_0)$, \ie, the common hinge loss with a bent slope at 0. The effect of such bent slope is to shrink $f$ to 0 when $p_1$ and $p_2$ are close. For $f^*(\bx)=\argmin_{f(\bx)}\E(\ell(Af)|\bX=\bx)$, we have $\phi^*=\sign(f^*)$.

The situation is much more complicated for multicategory classification. Suppose there are 3 classes, that is, $\mA=\{1, 2, 3\}$, then the possible values for the classifier $\phi(\bX)$ are $\{1\}$, $\{2\}$, $\{3\}$, $\{1, 2\}$, $\{1, 3\}$, $\{2, 3\}$, and $\{1, 2, 3\}$. In general, assuming there are $k$ classes, $\phi(\bX)$ can be any element in the power set of $\{1,\dots,k\}$ (except the empty set). In addition to the reject option, which can be written as $\{1,\dots,k\}$, \cite{zhang2018reject} introduced the so-called refine option, in which a refined decision has a cardinality greater than 1 and less than $k$. It contains all those class labels which are nearly as plausible as the most plausible class. \cite{zhang2018reject} proposed to use a class of loss functions in conjunction with the angle-based learning framework \cite{zhang2014multicategory} to train a set-valued classifier that can render these different options. We note that both the reject option and the refine option are set-valued decisions, and they are analogous to the set-valued recommendations in this article. 


\section{Methodology}\label{sec:method}
In this section, we introduce the framework of alternative individualized treatment recommendations (A-ITR) and propose two methods to estimate the optimal A-ITR.

\subsection{A-ITR Framework}
There are several situations in which ITRs with additional alternative options are desirable. Even with small errors, when several treatments are near the optimality, the ranking of these treatments based on their estimated outcomes may differ from their true ranking. In this case, reporting only one treatment based on the estimated value is problematic. Secondly, when the error in the learning problem is substantially large, the so-called optimal treatment reported by conventional ITRs may lead to an outcome that is much worse than some of the other treatment options. In these situations, recommending a single treatment only adds to the distrust that patients may already have towards such black-box algorithms that they know little about. On the other hand, A-ITR provides a safety net, preventing from committing to a single treatment that is only one out of multiple treatments with similar or indistinguishable outcomes. Morally, as patients are more mindful about their financial responsibility and their quality of life, it is more appropriate to present these alternative options and have the patients themselves to make an informed decision, especially when many of these decisions are life changing.

An A-ITR is a set-valued map $\phi: \mX\to2^\mA\backslash\emptyset$. Inspired by the idea of classification with reject and refine options, we formally define the optimal A-ITR as,
\begin{align}\label{opt_aitr}
\phi^*(\bx)=\{j;\mu_j(\bx)/\mu_{(1)}(\bx)\leq c\},
\end{align}
where $\mu_{(1)}<\dots<\mu_{(k)}$ denote the ordered conditional mean outcomes, and $c\geq1$ is a user-predefined number. This optimal A-ITR is a near-optimal treatment recommendation set since it contains all the treatment options with $\mu_j$ close to that of the optimal one, up to a multiplicative constant $c$. Note that $\phi^*$ may contain only one element, that is, the treatment with the smallest mean outcome, which corresponds to the conventional ITR. If it includes all the treatments, it is a non-informative recommendation, analogous to the reject option in set-valued classification. Here we call $c$ the near-optimal parameter.

\begin{figure}[!htb]
\begin{center}
\includegraphics[width=5.5in]{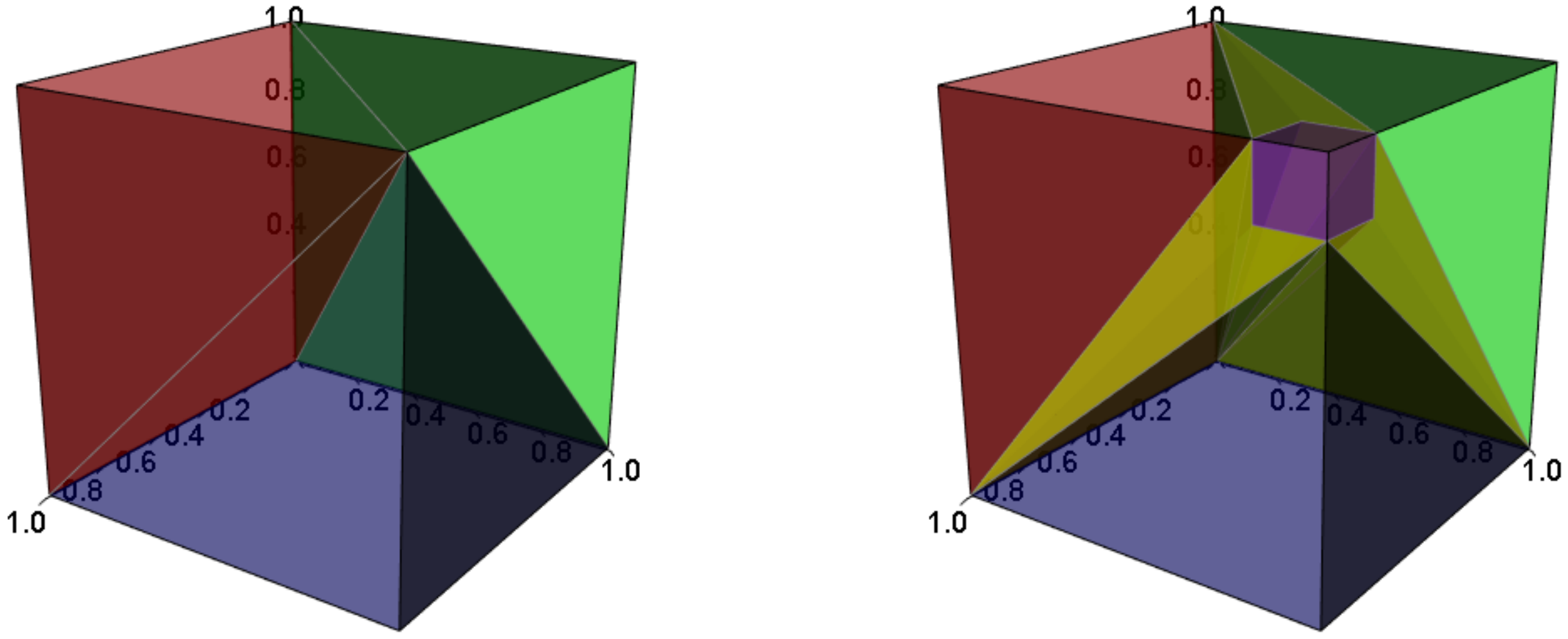}\vspace{-1.5em}
\end{center}
\caption{\small The optimal A-ITR when $k=3$ for different $c$ (left: $c=1$; right: $c=1.2$). Any point in the plot represents $(\mu_1, \mu_2, \mu_3)$ (suppose $Y^*(j)\in(0, 1)$) with the recommendation illustrated by colors. Points in the red, green, and blue regions contain only one treatment option; the yellow region contains recommendations with two options; and the purple region includes those with all three options.}\vspace{-1.5em}\label{fig:theoretical}
\end{figure}

When $c=1$, the optimal A-ITR reduces to the optimal ITR defined in (\ref{opt_itr}) since $\phi^*=\{j;\mu_j/\mu_{(1)}\leq1\}=\{j;\mu_j=\mu_{(1)}\}=\argmin_{j}\mu_j$. This means that the proposed optimal A-ITR is a generalization of the conventional optimal ITR. Figure \ref{fig:theoretical} shows theoretical regions for different types of recommendations from the optimal A-ITR for 2 different values of $c$ in a $k=3$ case.

\subsection{Estimation}\label{sec:est}
We consider two types of methods to estimate the optimal A-ITR: the regression-based methods and the classification-based methods. For regression-based methods, we can use Q-learning to first estimate the conditional mean $\mu_j(\bx)=\E(Y|\bx,A=j)$ for each treatment $j$, then plug into (\ref{opt_aitr}), \ie, $\hat\phi(\bx)=\{j;\hat\mu_j(\bx)/\hat\mu_{(1)}(\bx)\leq c\}$. The success of this regression-based plug-in method relies on accurate estimation of $\mu_j$.

In contrast, the classification-based method targets on estimating the true boundary between different decision regions, bypassing the need to estimate $\mu_j$ directly. In the rest of the section, we propose two classification-based methods within the OWL framework, both of which are based on the angle-based learning approach \citep{zhang2014multicategory}. 

\cite{zhang2018multicategory} first made use of the angle-based learning approach to solve the ITR problem, in which they denoted $\bW_1\dots,\bW_k$ as the vertices of a $(k-1)$-dimensional simplex and they chose the loss $L(A, \bdf)$ in (\ref{owl}) to be a function that only depends on the inner product $\<{\bW_A, \bdf}$, namely, $L(A, \bdf)=\ell(\<{\bW_A, \bdf})$. Define $\bdf^*$ to be the population minimizer under such loss, that is
\begin{align}\label{f^*}
\bdf^*=\argmin_{\bdf\in\{\mX\to\R^{k-1}\}}\E\left[\frac{Y}{p(A|\bX)}\ell(\<{\bW_A, \bdf(\bX)})\right].
\end{align}
The end product of \cite{zhang2018multicategory} was a single-treatment ITR. In the ideal case that $\bdf^*$ can be obtained, their ITR was defined as $d_{\bdf^*}(\bx)=\argmax_j\<{\bW_j, \bdf^*(\bx)}$, and it can be shown that as long as $\ell$ is convex and strictly increasing, Fisher consistency holds, \ie $d_{\bdf^*}=d^*$. In practice, given the training data set $\{(\bx_i, a_i, y_i)\}_{i=1}^n$, $\hat\bdf$, the estimate of $\bdf^*$, is obtained by,
\begin{align}\label{fhat}
\hat\bdf=\argmin_{\bdf\in\mF}\frac{1}{n}\sum_{i=1}^n\frac{y_i}{p(a_i|\bx_i)}\ell(\<{\bW_{a_i}, \bdf(\bx_i)}),\ \text{subject to}\ J(\bdf)\leq s,
\end{align}
where $\mF\subseteq\{\bdf;\mX\to\R^{k-1}\}$ is a class of functions, and $J(\bdf)$ is a penalty term to prevent overfitting. 

Both our proposed A-ITR methods are derived from the population minimizer $\bdf^*$ (\ref{f^*}) or the empirical minimizer $\hat \bdf$ (\ref{fhat}) obtained in this way. The differences lie in the loss function $\ell$ they use, and how they convert $\bdf^*$ or $\hat \bdf$ to the final set-valued recommendations.
\subsubsection{Two-step OWL Method}
For the two-step method, we use a convex, differentiable, and increasing loss function $\ell_D$. To calculate the A-ITR, we have to first order the vertices, for any $\bdf$ (which may be $\bdf^*$ or $\hat \bdf$), in the manner of order statistics, \ie, $\<{\bW_{(1)}, \bdf}>\dots>\<{\bW_{(k)}, \bdf}$. The resultant two-step estimator of the optimal A-ITR is then defined as
\begin{align}\label{phi^D}
\phi_\bdf^D(\bx)=\left\{j;\frac{\ell'_D(\<{\bW_{(1)}, \bdf(\bx)})}{\ell'_D(\<{\bW_j, \bdf(\bx)})}\leq c\right\}.
\end{align}
Here $\ell'_D$ is the first derivative of $\ell_D$, and the superscript ``$D$" indicates that $\bdf$ is the population solution to (\ref{f^*}) or the empirical solution to (\ref{fhat}), based on a \textit{differentiable} loss function. Our estimator is partially justified and motivated by the following result.
\begin{proposition}[\citealp{zhang2018multicategory}]\label{prop:diff}
Let $\bdf^*$ be the population minimizer in (\ref{f^*}) in which $\ell$ is a convex and differentiable function $\ell_D$ with $\ell'_D(u)>0$ for all $u$. For any $i\neq j\in\{1,\dots,k\}$, we have
$$\frac{\mu_j}{\mu_i}=\frac{\ell'_D(\<{\bW_i, \bdf^*})}{\ell'_D(\<{\bW_j, \bdf^*})}.$$
\end{proposition}

Proposition \ref{prop:diff} implies the following Fisher-consistent-like result for our proposed A-ITR estimator.

\begin{proposition}\label{prop:two_step}
Let $\bdf^*$ be the population minimizer in (\ref{f^*}) in which $\ell$ is a convex and differentiable function $\ell_D$ with $\ell'_D(u)>0$ for all $u$. The A-ITR $\phi_{\bdf^*}^D$ defined in (\ref{phi^D}) based on $\bdf^*$ coincides with the optimal A-ITR $\phi^*$ (\ref{opt_aitr}).
\end{proposition}

This method is a two-step procedure because it first estimates the ratios of conditional means ${\mu_j}/{\mu_i}$ using ${\ell'_D(\<{\bW_i, \hat\bdf})}/{\ell'_D(\<{\bW_j, \hat\bdf})}$ then plugs it back into (\ref{opt_aitr}). Note that it does not estimate each conditional mean individually, but their ratios. The issue remains that if $\bdf^*$ cannot be accurately estimated, the ratio of the conditional means cannot be accurately estimated.

\subsubsection{One-step OWL Method}\label{sec:one_step}
\begin{figure}[!htb]
\begin{center}
\includegraphics[width=5in]{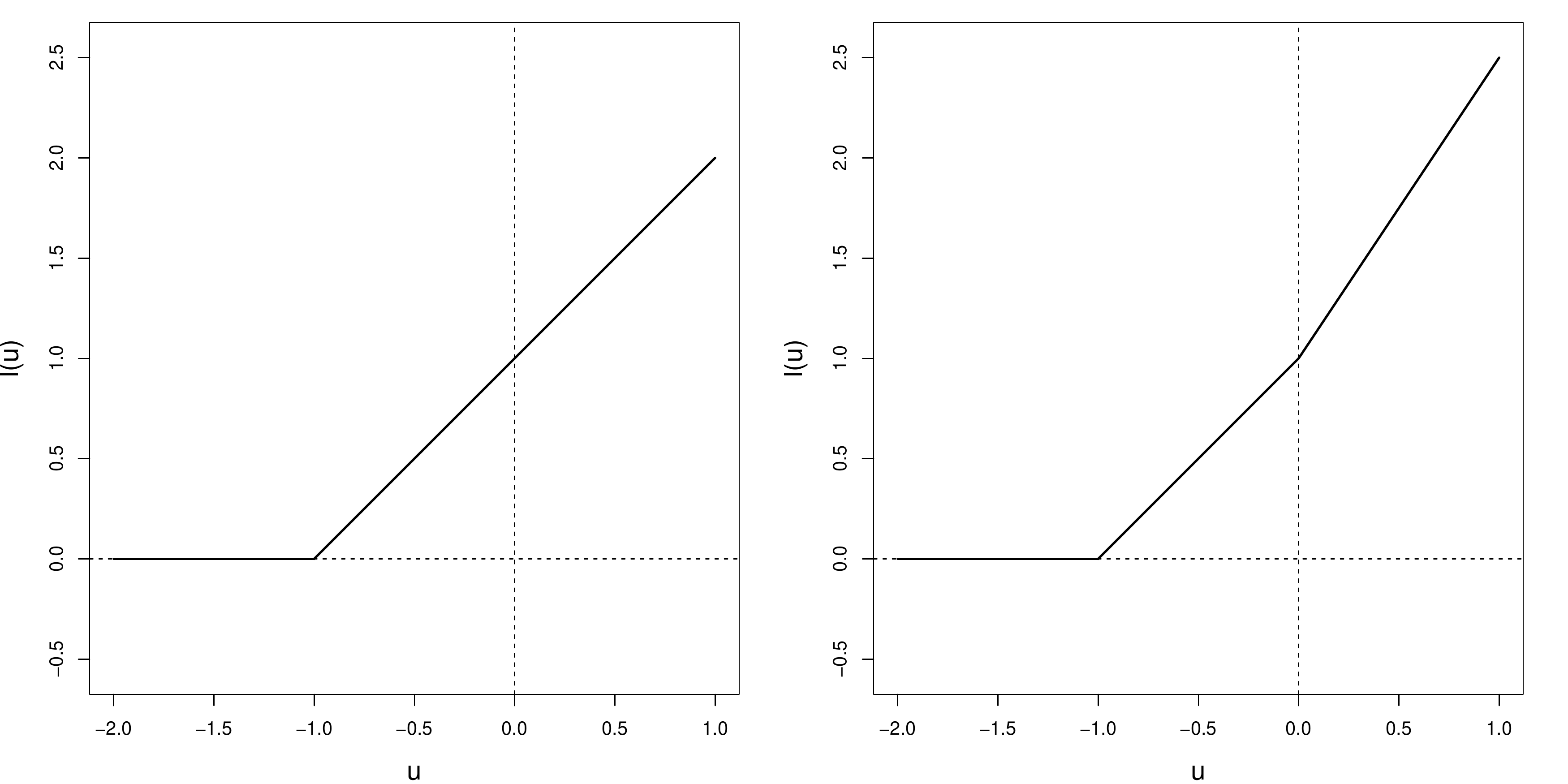}\vspace{-1.5em}
\end{center}
\caption{\small Hinge loss (left panel) and bent hinge loss with $c=1.5$ (right panel).}\label{fig:loss}
\end{figure}

The one-step method aims to directly obtain a set-valued recommendation without calculating ${\ell'_D(\<{\bW_i, \hat\bdf})}/{\ell'_D(\<{\bW_j, \hat\bdf})}$. The crucial difference here is the use of a bent loss function, defined as
$$\ell_B(\<{\bW_j, \bdf})=\ell_1(\<{\bW_j, \bdf})+\ell_2(\<{\bW_j, \bdf}),$$
where $\ell_1>0$ is a convex and increasing function with $\ell_1'(u)=1$ for all $u\geq0$, and $\ell_2(u)=(c-1)u_+$ with $c\geq1$. Such a loss function is bent at 0, since $\ell_B'(0-)=1$ and $\ell_B'(0+)=c$. An example of bent loss is the bent hinge loss, $\ell_B(u)=(1+u)_++(c-1)u_+$ (see Figure \ref{fig:loss}.) The bent loss has been a critical tool that helps to achieve reject (and refine) options in the classification literature \citep{bartlett2008classification, zhang2018reject}. 

The main effect of the bent loss is to shrink the so-called angle margins for class (or treatment) $j$, defined as $\<{\bW_j, \bdf}$, towards 0, similar to the shrinkage effect of the lasso penalty. Likewise, the additional slope $c\geq1$, the near-optimal parameter, for positive $u$ is analogous to a larger penalty parameter in lasso regression, which would encourage a sparser model. Note that here such a shrinkage effect is applied to the classes (treatments) with nonnegative angle margins only. Specifically, the angle margin for class $j$ will tend to be shrunken to 0 if $1<\mu_j/\mu_{(1)}\leq c$.
Formally, $\bdf^*$, the solution to (\ref{f^*}) with the bent loss $\ell_B$, enjoys the following theoretical property. Proposition \ref{prop:bent} is derived from Proposition 1 in \cite{zhang2018reject}

\begin{proposition}\label{prop:bent}
For the sequence $\mu_{(1)}<\dots<\mu_{(k)}$, if there exists an integer $r\in\{1,\dots,k-1\}$ such that $\mu_{(j)}/\mu_{(1)}<c$ for $j=1,\dots,r$ and $\mu_{(j)}/\mu_{(1)}>c$ for $j=r+1,\dots,k$, then $\bdf^*$ as defined in (\ref{f^*}) in which $\ell$ is a convex and increasing function $\ell_B$ with $\ell_B'(0-)=1$ and $\ell_B'(0+)=c\geq1$ satisfies that $\<{\bW_{(1)}, \bdf^*}>0$, $\<{\bW_{(2)}, \bdf^*}=\dots=\<{\bW_{(r)}, \bdf^*}=0$, and $\<{\bW_{(r+1)}, \bdf^*}=\dots=\<{\bW_{(k)}, \bdf^*}<0$; otherwise, $\<{\bW_{(j)}, \bdf^*}=0$ for $j=1,\dots,k$.
\end{proposition}

Proposition \ref{prop:bent} shows that when a bent loss is used, the population minimizer $\bdf^*$ (\ref{f^*}) has the nice property that all those treatments within the near-optimal net (defined as $\mu_{(j)}/\mu_{(1)}\leq c$) have non-negative angle margins; all those treatments outside the near-optimal net have negative angle margins. This naturally leads to the following set-valued recommendation,
\begin{align}\label{phi^B}
\phi_\bdf^B(\bx)=\{j;\<{\bW_j, \bdf(\bx)}\geq0\}.
\end{align}
The superscript ``$B$" means that $\bdf$ is the population minimizer $\bdf^*$ (\ref{f^*}) or the empirical minimizer $\hat \bdf$ (\ref{fhat}), with the loss $\ell$ being a \textit{bent} loss $\ell_B$, as opposed to a differentiable loss function in the two-step method.

Proposition \ref{prop:bent} implies that $\{j;\mu_j(\bx)/\mu_{(1)}(\bx)<c\}\subseteq\phi_{\bdf^*}^B(\bx)\subseteq\{j;\mu_j(\bx)/\mu_{(1)}(\bx)\leq c\}$. The following assumption is necessary to resolve the identifiable issue of (\ref{phi^B}) and to show its optimality.

\begin{assumption}\label{asp:0measure}
For any positive $c_0$, $P\left(\mu_j(\bX)=c_0\mu_i(\bX)\right)=0$ for $\forall i\neq j\in\{1,\dots,k\}$ in which $\mu_j(\bX)=\E(Y|\bX,A=j)$ is the conditional mean outcome for treatment $j$.
\end{assumption}

Assumption \ref{asp:0measure} guarantees the sets $\{\bx\in\mX;\mu_j(\bx)/\mu_i(\bx)=1\}$ and $\{\bx\in\mX;\mu_j(\bx)/\mu_i(\bx)=c\}$ (where $c$ is the near-optimal parameter in (\ref{opt_aitr})) have measure 0 for any $i\neq j$ so that $\bdf^*$ is identifiable almost everywhere. Under Assumption \ref{asp:0measure}, we have the following proposition, analogous to Fisher consistency in classification.

\begin{proposition}\label{prop:one_step}
Suppose Assumption \ref{asp:0measure} holds. Let $\bdf^*$ be the population minimizer in (\ref{f^*}) in which $\ell$ is a convex and increasing function $\ell_B$ with $\ell_B'(0-)=1$ and $\ell_B'(0+)=c\geq1$. The A-ITR $\phi_{\bdf^*}^B$ defined in (\ref{phi^B}) based on $\bdf^*$ coincides with the optimal A-ITR $\phi^*$ (\ref{opt_aitr}) with the near-parameter parameter $c$.
\end{proposition}

Note that for both classification-based methods, a single-valued ITR can be easily defined by recommending the treatment option with the largest angle margin, that is, $\argmax_j\<{\bW_j, \bdf(\bx)}$.

Unlike the regression-based method, the two classification-based methods do not estimate the conditional mean outcome. The success of the regression-based method relies on accurate estimation of $\mu_j$ at every $\bx$ of interest, while reasonable performance is expected for the classification-based methods as long as the estimation is accurate around the ``boundaries". However, the two-step method and the one-step method seem to have different focuses. Both methods start with finding a discriminant function to minimize outcome-weighted classification error for the purpose of minimizing the expected outcome. As a consequence, both methods have ``good'' performances near boundaries that distinguish the optimal treatment from the non-optimal treatments for each patient. The one-step method, additionally, uses a bent loss with a shrinkage effect that is capable of determining whether a treatment is {\it close enough to}, not whether it is {\it equal to}, the optimal treatment (more precisely speaking, whether the ratio between the conditional outcomes $\mu_j(\bx)/\mu_{(1)}(\bx)$ is less than $c$ or not). This is theoretically justified by Proposition \ref{prop:bent}. Hence, 
the one-step method also has ``good'' performance near such new notions of boundaries (that is, boundaries between the top few ``equally good'' treatment options, and the others.)

\begin{figure}[!htb]
\begin{center}
\includegraphics[width=\textwidth]{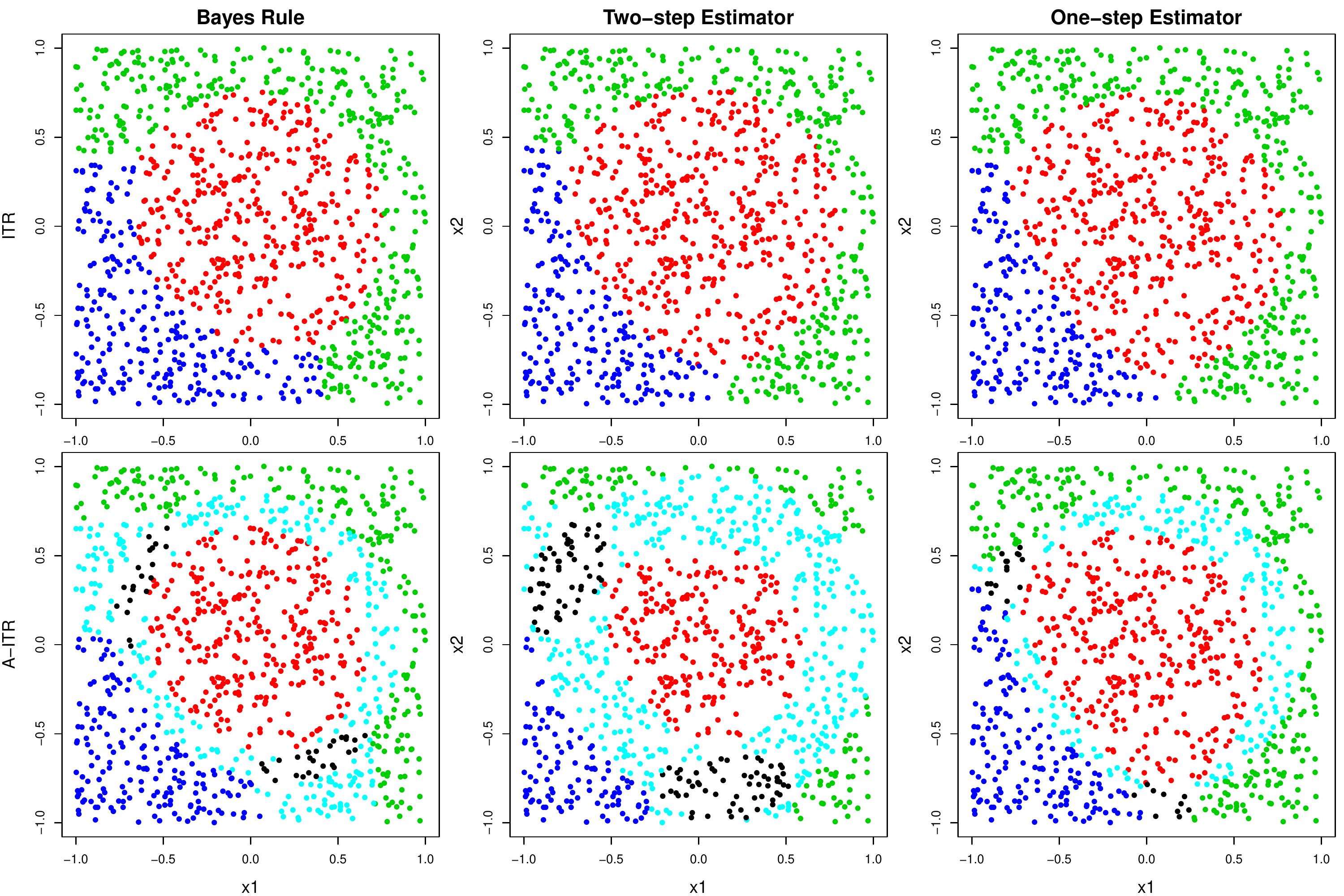}\vspace{-1.5em}
\end{center}
\caption{\small ITR (top) and A-ITR (bottom) for a toy example given by Bayes rule, the two-step estimator and one-step estimator. Cyan color indicates recommendations with two options, and black color indicates three options. Both estimators give similar results to the Bayes rule in terms of the single-valued ITR. The two-step method does not provide as good A-ITR results as the one-step method.}\vspace{-1em}\label{fig:toy}
\end{figure}

To illustrate the additional strength of the one-step method, we show the boundaries between recommendations for a toy example (the details of which will be revisited in the numerical studies) in Figure \ref{fig:toy}, in which the top row shows the single-valued ITR and the second row the set-valued A-ITR, by the Bayes rule, the two-step method and the one-step method respectively. Both classification-based methods give good approximations to the Bayes ITR boundaries, shown in the top row. However, the two-step method seems to have a bigger error in terms of the A-ITR boundaries when compared to the Bayes rule (shown in the bottom row), than the one-step method does. This is probably due to the fact that the optimization for the two-step method is not designed to capture this subtle pattern, at least not for a finite sample.

A potential drawback of the one-step method is that it relies on an additional Assumption \ref{asp:0measure} to be identifiable. In general such an assumption is very weak, though it can be nontrivial in cases where the outcome can only take finite values. For example, if the possible outcomes are integers between 1 to 10, then Assumption \ref{asp:0measure} may not hold for some $c$.

\section{Implementations}\label{sec:opt}
In this section, we discuss various aspects of the implementations for the proposed methods, including the optimization, the normalization of the predictive function, and the parameter tuning. 

\subsection{Algorithm}
In this section, we introduce the optimization procedure to estimate $\bdf^*$ defined in (\ref{f^*}). Instead of the constrained problem (\ref{fhat}), we solve the regularized problem:
\begin{align}\label{optimization}
\min_{\bdf\in\mF}\frac{1}{n}\sum_{i=1}^n\frac{y_i}{p(a_i|\bx_i)}\ell(\<{\bW_{a_i}, \bdf(\bx_i)})+\lambda J(\bdf),
\end{align}
where $\lambda$ is a tuning parameter. It is a weighted classification problem with weight $w_i=y_i/p(a_i|\bx_i)$.

In terms of the function class $\mF$, there are linear learning and kernel learning \citep{steinwart2007fast, hofmann2008kernel, hastie2009elements}. Let $\bdf=(f_1,\dots,f_{k-1})^T\in\mF$, and for simplicity, we add a constant term to $\bx$. Then for linear learning, we have $f_j(\bx)=\bx^T\bbeta_j$, and the corresponding penalty $J(\bdf)=\sum_{j=1}^{k-1}||\bbeta_j||^2=\sum_{j=1}^{k-1}\bbeta_j^T\bbeta_j$. For kernel learning, $f_j(\bx)=\sum_{i=1}^nK(\bx_i, \bx)\alpha_{ij}+\alpha_{0j}$, where $K(\cdot, \cdot)$ is a kernel function. The penalty term becomes $J(\bdf)=\sum_{j=1}^{k-1}\balpha_j^TP\balpha_j+\sum_{j=1}^{k-1}\alpha_{0j}^2$, where $P$ is the gram matrix. Note that we include the intercept term into $J(\bdf)$ and a benefit by doing this is the reduction of the complexity of the algorithm. \cite{zhang2016quantile} shows theoretically that it can achieve the same convergence rate as the case without the intercept term.

We proposed the two-step method and the one-step method. The two-step method is based on a differentiable loss $\ell_D$, while the one-step method is based on a bent loss $\ell_B(u)=\ell_1(u)+\ell_2(u)$, where $\ell_1$ is convex and $\ell_2(u)=(c-1)u_+$. Since $\ell_D$ is similar to a special case of $\ell_B$ with $c=1$, here we only need to focus on the algorithm for the bent loss $\ell_B$. In the rest of this section, we use linear learning to demonstrate our algorithm, and have deferred the details about kernel learning to the supplementary materials.

We first consider the case when $\ell_1$ is differentiable. In this case, we use the ADMM \citep{boyd2011distributed} algorithm to solve (\ref{optimization}). The ADMM algorithm is used when the objective function can be written as a sum of two convex functions, which, in our case, are $\ell_1$ and $\ell_2$.

To start with, we denote the coefficient matrix as $B_{p\times(k-1)}=[\bbeta_1,\dots,\bbeta_{k-1}]$. Then we create another copy of the coefficients $G_{p\times(k-1)}=[\bgamma_1,\dots,\bgamma_{k-1}]$, and let $Z_{p\times(k-1)}=[\bz_1,\dots,\bz_{k-1}]$. Recall that $w_i=y_i/p(a_i|\bx_i)$, then we minimize the augmented Lagrangian
\begin{align*}
L_\rho(B, G, Z)=&\sum_{i=1}^nw_i\ell_1(\<{\bW_{a_i}, B^T\bx_i})+\sum_{i=1}^nw_i\ell_2(\<{\bW_{a_i}, G^T\bx_i})+\frac{n\lambda}{2}\sum_{j=1}^{k-1}\bbeta_j^T\bbeta_j\\
&+\sum_{j=1}^{k-1}\bz_j^T(\bbeta_j-\bgamma_j)+\frac{\rho}{2}\sum_{j=1}^{k-1}(\bbeta_j-\bgamma_j)^T(\bbeta_j-\bgamma_j),
\end{align*}
where $\rho>0$ controls the step size.

At step $t$, for each $j=1,\dots,k-1$ we can update $B^t$, $G^t$ and $Z^t$ as
\begin{align*}
&\bbeta_j^t=\argmin_{\bbeta_j}L_\rho([\bbeta_1^t,\dots,\bbeta_j,\dots,\bbeta_{k-1}^{t-1}], G^{t-1}, Z^{t-1}),\\
&\bgamma_j^t=\argmin_{\bgamma_j}L_\rho(B^t, [\bgamma_1^t,\dots,\bgamma_j,\dots,\bgamma_{k-1}^{t-1}], Z^{t-1}),\\
&\bz_j^t=\bz_j^{t-1}+\rho(\bbeta_j^t-\bgamma_j^t)
\end{align*}
until matrix $B$ converges. Note that in the two-step method where $c=1$, we have $\ell_2(u)=0$. In this case, we can force $B=G$ and only update $\bbeta_j^t$'s until they converge.

Next we consider the case when $\ell_1$ is not differentiable. In the literature of classification, a non-differentiable loss that has been commonly used is hinge loss. Note that in our case, since we prefer smaller outcomes, we define the hinge loss as $\ell_1(u)=(1+u)_+$ (see Figure \ref{fig:loss}). That is, we flip the traditional hinge loss around the y-axis to make it an increasing function. A typical approach to a problem with the hinge loss is to transform it into a quadratic programming (QP) problem in its duality \citep{fung2005multicategory, hastie2009elements, zhang2018reject}. Specifically, the dual problem of (\ref{optimization}) can be written as
\begin{align*}
\min_{\alpha_j,\gamma_j}\ &\frac{n\lambda}{2}\sum_{j=1}^{k-1}\bbeta_j^T\bbeta_j-\sum_{i=1}^n\alpha_i\\
\text{s.t. }&0\leq\alpha_i\leq w_i,\ 0\leq\gamma_i\leq w_i,\ i=1,\dots,n,
\end{align*}
where $\bbeta_j=-\frac{1}{n\lambda}\sum_{i=1}^n(\alpha_i+(c-1)\gamma_i)W_{a_i,j}\bx_i$, and $W_{a_i,j}$ is the $j$th component of $\bW_{a_i}$. Note that the weight $w_i=y_i/p(a_i|\bx_i)$ serves as the upper bound of the box constraints. Because the objective function is quadratic in $\alpha_i$ and $\gamma_i$, it has explicit solution at each iteration. Thus it converges very fast by using algorithms such as coordinate decent \citep{zhang2018reject}.

In practice there may be numerical errors to the solution. Moreover, due to different choices of the tuning parameter $\lambda$, the scale of the resulting angle margins may vary much between different tuning trials. We propose the following normalization procedure for the one-step A-ITR $\phi_\bdf^B$ (\ref{phi^B}) to boost the empirical performance. The idea is that instead of recommending all treatments with angle margins greater than or equal to 0, we change the threshold to a small number varying around 0. Such a threshold is a fixed constant $\delta$ multiplied by a measure of the scale, chosen to be the magnitude of the smallest angle margin. The normalized two-step A-ITR is then
\begin{align}\label{phihat}
\phi_{\hat\bdf}^B(\bx)=\{j;\<{\bW_j, \hat\bdf(\bx)}\geq\delta M(\bx)\},
\end{align}
where $\delta$ is a tuning parameter around 0 and $M(\bx)=|\<{\bW_{(k)}, \hat\bdf(\bx)}|$ is the magnitude of the smallest angle margin (note that $\<{\bW_{(k)}, \hat\bdf(\bx)}$ is negative).

\subsection{Tuning Procedure}\label{sec:tune}
In this paper, the estimation procedure involves two tuning parameters. The first one is the regularization parameter $\lambda$ in (\ref{optimization}) which appears in both the two-step and one-step methods. The second one is the normalization parameter $\delta$ in (\ref{phihat}) for the one-step method only. We will tune these two parameters differently in two steps.

The first step is to tune $\lambda$. For each $\lambda$, the estimated solution is $\hat\bdf$. Then we define the corresponding single-treatment ITR as $d_{\hat\bdf}=\argmax_j\<{\bW_j, \hat\bdf}$ and calculate its empirical average of the expected outcome (\ref{E^d}), which is given by $$\frac{\sum_{i=1}^n\left(\ind{a_i=d_{\hat\bdf}(\bx_i)}y_i/p(a_i|\bx_i)\right)}{\sum_{i=1}^n\left(\ind{a_i=d_{\hat\bdf}(\bx_i)}/p(a_i|\bx_i)\right)}$$
\citep{zhao2012estimating, zhang2018multicategory}. We choose the $\lambda$ that yields the smallest empirical risk for the resulting ITR, even if our ultimate goal is to obtain a set-valued A-ITR. This can substantially simplify the tuning process. We found that other more complicated tuning procedures have led to a similar performance. 

For the one-step method, we need to continue to tune $\delta$. For the same $\lambda$ (same resulting ITR), because different $\delta$'s may lead to slightly different set-valued A-ITRs and recommendations with different carnalities, we must actually compare the resulting A-ITRs to choose the best $\delta$, instead of using the ITR as a proxy. However, there are some difficulties in evaluating the performance of the estimated A-ITR. Compared to the conventional ITR, the challenge here is that when the recommendation includes two or more treatment options, there are multiple potential outcomes and it is difficult to quantify the ``overall" benefit for such a recommendation.

Although the proposed optimal A-ITR $\phi^*$ defined in (\ref{opt_aitr}) is not a Bayes rule under any loss function, we can consider a closely related loss function, whose risk function is given by
\begin{align}\label{wt_loss}
\E\left[\frac{Y\ind{A\in\phi(\bX)}}{p(A|\bX)(1+(|\phi(\bX)|-1)c)}\right],
\end{align}
where $\phi: \mX\to2^\mA\backslash\emptyset$ is a set-valued predictor and $|\cdot|$ denotes the cardinality of a set. Compared to the expected outcome $\E^d(Y)$ defined in (\ref{E^d}), this quantity is a weighted outcome with weight $1/(1+(|\phi|-1)c)$ under $\phi$. If we force $|\phi|\equiv 1$, it reduces to $\E^d(Y)$. 
More importantly, it can be shown that the minimizer of (\ref{wt_loss}), denoted by $\phi^+$, is
$$\phi^+(\bx)=\argmin_{\phi(\bx)\in 2^\mA}\mu^\phi(\bx),\quad\text{where}\ \mu^{\phi}(\bx)\triangleq\frac{1}{1+(|\phi(\bx)|-1)c}\sum_{j\in\phi(\bx)}\mu_j(\bx).$$
Here $\mu^\phi$ defines a new criterion that generalizes the expected outcomes under a set-valued treatment recommendation $\phi$. To see that, note that for $\phi(\bx)=\{1\}$, $\mu^\phi(\bx)=\mu_1(\bx)$, while for $\phi(\bx)=\{1, 2\}$, $\mu^\phi(\bx)=(\mu_1(\bx)+\mu_2(\bx))/(1+c)$, which is smaller than the simple average $(\mu_1(\bx)+\mu_2(\bx))/2$ when $c>1$. Suppose treatment 1 is better than treatment 2 ($\mu_1(\bx)<\mu_2(\bx)$). We can show that $\phi_2(\bx)\triangleq\{1, 2\}$ is as good as $\phi_1(\bx)\triangleq\{1\}$ under this new criterion if and only if $\mu_2(\bx)/\mu_1(\bx)\leq c$, which is exactly the near-optimal recommendation set defined in (\ref{opt_aitr}). 

\begin{figure}[!htb]
\begin{center}
\includegraphics[width=5.5in]{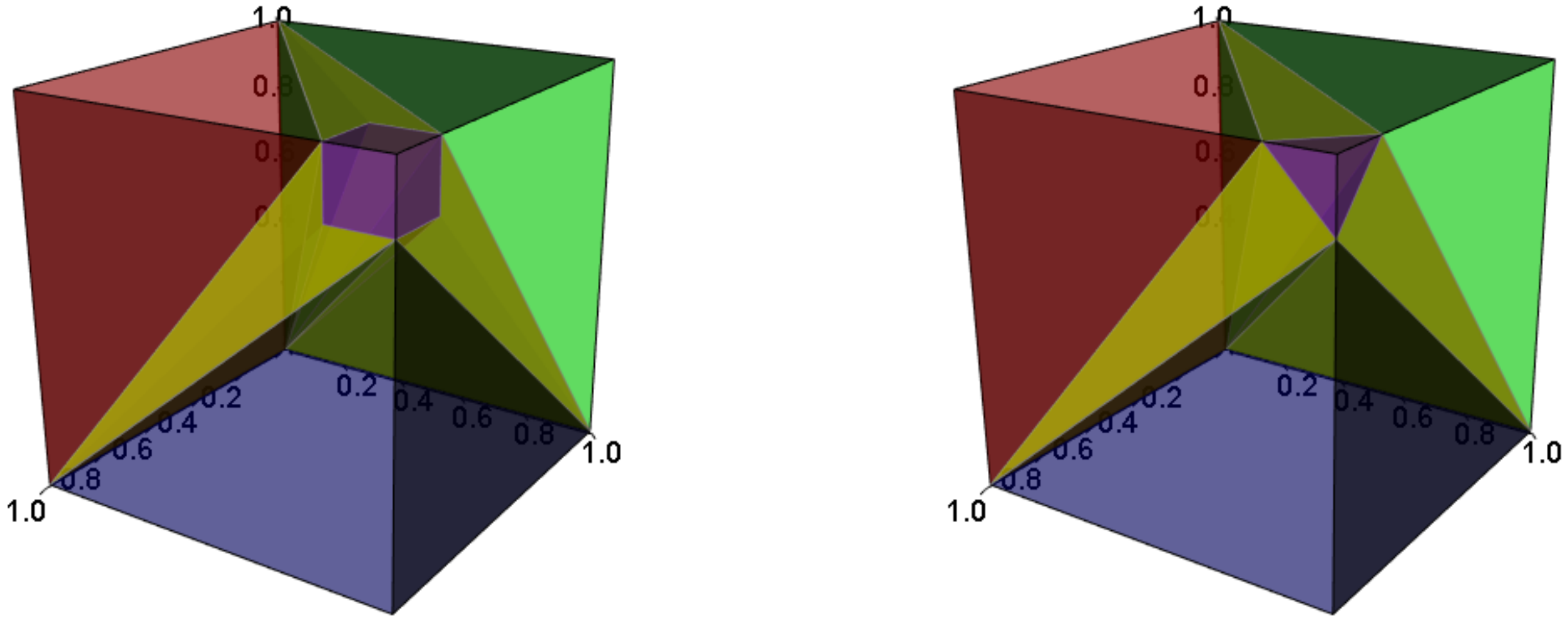}\vspace{-1.5em}
\end{center}
\caption{\small Comparison between $\phi^*$ (left panel) and $\phi^+$ (right panel) when $k=3$ and $c=1.2$. Any point in the plot represents $(\mu_1, \mu_2, \mu_3)$ (suppose $Y^*(j)\in(0, 1)$) with the recommendation illustrated by colors. As in Figure \ref{fig:theoretical}, the red, green, and blue regions contain only one treatment; the yellow region contains two treatments; and the purple region includes all three treatments. $S_1^*=S_1^+$ (unions of red, blue and green regions), and $S_2^*\subseteq S_2^+$ (all but the purple regions).}\vspace{-1em}\label{fig:two_rules}
\end{figure}

Intuitively, $\phi^+(\bx)$ is an optimal set of treatments selected to minimize the ``average" clinical outcome with a penalty on the cardinality of the recommendation set. Note that when $c=1$, $\phi^+$ is the same as the optimal ITR $d^*$. Moreover, when $k=2$, $\phi^+$ is the same as the optimal A-ITR $\phi^*$, as shown above. When $k\geq3$, $\phi^+$ and $\phi^*$ are different but are nested within each other in the following way: if we let $S_t^*=\{\bx\in\mX;|\phi^*(\bx)|\leq t\}$ and $S_t^+=\{\bx\in\mX;|\phi^+(\bx)|\leq t\}$, then we have $S_1^*=S_1^+$, and $S_t^*\subseteq S_t^+$ for $t=2,\dots,k-1$. Figure \ref{fig:two_rules} demonstrates their relationship when $k=3$.

From Figure \ref{fig:two_rules}, we observe that the regions with only one treatment are the same ($S_1^*=S_1^+$), while the regions containing two or three treatments are slightly different. In general, the boundaries between the size-1 decisions and their complements are the same for the two rules $\phi^+$ and $\phi^*$. They only differ in the boundaries between recommendations with different cardinalities (for example, the boundary between size-2 decisions and size-3 decisions). Due to the similarity between $\phi^+$ and $\phi^*$, although $\phi^*$ does not minimize the weighted outcome defined in (\ref{wt_loss}), it is close. This justifies the use of the weighted outcome (\ref{wt_loss}) as a new criterion for the tuning parameter selection. Specifically, we choose the $\delta$ value that can yield the smallest value of the following empirical counterpart of (\ref{wt_loss}),
\begin{align}\label{empr_wt_loss}
\frac{\sum_{i=1}^n\left(\ind{a_i\in\phi_{\hat\bdf}(\bx_i)}y_i/\left[p(a_i|\bx_i)(1+(|\phi_{\hat\bdf}(\bx_i)|-1)c)\right]\right)}{\sum_{i=1}^n\left(\ind{a_i\in\phi_{\hat\bdf}(\bx_i)}/\left[p(a_i|\bx_i)|\phi_{\hat\bdf}(\bx_i)|\right]\right)}.
\end{align}
In addition to the tuning parameter selection, we may also use this criterion to select different methods for conducting A-ITRs. In the real data analysis, we will use this criterion to select between the two proposed classification-based methods.

\section{Statistical Learning Theory}\label{sec:theory}
In this section, we study the convergence rate of the excess $\ell$-risk in both linear learning and kernel learning settings. We assume the random vector $\bZ=(\bX, A, Y)$ follows a certain distribution $P$ that satisfies Assumption \ref{asp:reg}. Furthermore, we make an additional assumption.

\begin{assumption}\label{asp:wt_bd}
There is a constant $C>0$ such that $|Y/p(A|\bX)|\leq C$ holds. For simplicity, we set $C=1$ through out this section.
\end{assumption}

For $\bdf$ and $\bdf'$, two $(k-1)$-dimensional functions, and $\ell$, an increasing, convex and Lipchitz loss function, denote
$$e_\ell(\bdf, \bdf')=\E\left[\frac{Y}{p(A|\bX)}\ell(\<{\bW_A, \bdf})\right]-\E\left[\frac{Y}{p(A|\bX)}\ell(\<{\bW_A, \bdf'})\right].$$ We call $e_\ell(\bdf, \bdf^*)$ the excess $\ell$-risk of $\bdf$ if $\bdf^*$ is optimal within a certain function space $\mF$.

\subsection{Linear Learning}
We first consider the linear function space, that is, we assume $\bdf=(f_1,\dots,f_{k-1})^T$ with $f_j(\bx)=\bx^T\bbeta_j$ for $j=1,\dots,k-1$. For simplicity, we assume each covariate is bounded by $[0, 1]$.

\begin{assumption}\label{asp:x_bd}
$\bX\in\mX=[0, 1]^p$.
\end{assumption}

Now consider the following function space,
$$\mF(p, s)=\{\bdf=(f_1,\dots,f_{k-1})^T;f_j(\bx)=\bx^T\bbeta_j, j=1,\dots,k-1, J(\bdf)\leq s\},$$
where $J(\bdf)=\sum_{j=1}^{k-1}||\bbeta_j||_2^2=\sum_{j=1}^{k-1}\sum_{l=1}^p\beta_{lj}^2$. Let $\mF(p)=\cup_{0\leq s<\infty}\mF(p, s)$. Define
\begin{align*}
\bdf^{(p)}&=\argmin_{\bdf\in\mF(p)}\E\left[\frac{Y}{p(A|\bX)}\ell(\<{\bW_A, \bdf})\right],\\
\bdf^{(p, s)}&=\argmin_{\bdf\in\mF(p, s)}\E\left[\frac{Y}{p(A|\bX)}\ell(\<{\bW_A, \bdf})\right],\quad\text{and}\\
\hat\bdf&=\argmin_{\bdf\in\mF(p, s)}\frac{1}{n}\sum_{i=1}^n\frac{y_i}{p(a_i|\bx_i)}\ell(\<{\bW_{a_i}, \bdf(\bx_i)}).
\end{align*}
Theorem \ref{thm:lin} gives the convergence rate for the excess $\ell$-risk $e_\ell(\hat\bdf, \bdf^{(p)})$, where $p=p_n$, $s=s_n$ can grow with $n$ as $n\to\infty$.

\begin{theorem}\label{thm:lin}
Let $\tau_n=(n^{-1}\ln p_n)^{1/2}\to0$ as $n\to\infty$. For linear learning, suppose Assumptions \ref{asp:reg}, \ref{asp:0measure}, \ref{asp:wt_bd}, and \ref{asp:x_bd} hold. We have
$$e_\ell(\hat\bdf, \bdf^{(p_n)})=O\left(\max(c(p_ns_n)^{1/2}\tau_n\ln\tau_n^{-1}, \delta_n)\right),$$
almost surely under $P$, where $\delta_n=e_\ell(\bdf^{(p_n, s_n)}, \bdf^{(p_n)})$.
\end{theorem}

In Theorem \ref{thm:lin}, $\delta_n$ stands for the approximation error between the optimal $\bdf$ in $\mF(p_n, s_n)$ and the optimal $\bdf$ in $\mF(p_n)$. So if $s_n\to\infty$, $\delta_n$ converges to 0. On the other hand, the first term $O\left(c(p_ns_n)^{1/2}\tau_n\ln\tau_n^{-1}\right)$ is the estimation error between $\hat\bdf$ and $\bdf^{(p_n, s_n)}$, and as we increase $s_n$, it becomes larger. The optimal tuning parameter $s_n$ is then chosen such that $c(p_ns_n)^{1/2}\tau_n\ln\tau_n^{-1}\sim\delta_n$.

In Theorem \ref{thm:lin} we may allow $s_n\to\infty$ with an appropriately chosen rate. The reason is that when we include diverging number of covariates, \ie, $p\to\infty$, $\bdf^{(p)}$ can become more complicated, and thus we need a larger $s_n$ to accommodate this change. However, in practice it may not be necessary since the true model usually depends on a finite number of covariates. So we could simplify Theorem \ref{thm:lin} if we make the assumption that there is a finite $s^*$ such that $\bdf^{(p)}\in\mF(p, s^*)$ for all $p$. For example, suppose $f^*_j(\bx)=f^{(p)}_j(\bx)=\sum_{l=1}^mx_l\beta^*_{lj}$ for $j=1,\dots,k-1$ and all $p=m,\dots,\infty$. Then we can choose $s^*=\sum_{j=1}^{k-1}\sum_{l=1}^m|\beta^*_{lj}|^2$.

\begin{corollary}\label{cor:lin}
Suppose $\bdf^*$ defined in (\ref{f^*}) only depends on finite many covariates, and that Assumptions \ref{asp:reg}, \ref{asp:0measure}, \ref{asp:wt_bd}, and \ref{asp:x_bd} hold. We have $$e_\ell(\hat\bdf, \bdf^*) = O\left(cp_n^{1/2}\tau_n\ln\tau_n^{-1}\right)=O\left(c(n^{-1}p_n\ln p_n)^{1/2}\ln(n(\ln p_n)^{-1})\right),$$
almost surely under $P$.
\end{corollary}

The convergence of excess $\ell$-risk $e_\ell(\hat\bdf, \bdf^*)$ in Corollary \ref{cor:lin} requires that $p_n=o(n)$. Particularly, when $p_n$ grows no faster than $n^{1-r}$, where $0<r<1$, it can be verified that the error rate is at an order of no greater than $n^{-r/2}(\ln n)^{3/2}$. This result is consistent with most of the classical asymptotic theory that the dimension of covariates should not be greater than the number of observations. Furthermore, we observe that if $p_n=O(1)$, then $e_\ell(\hat\bdf, \bdf^*)=O(n^{-1/2}\ln n)$, which is almost $O(n^{-1/2})$.

\subsection{Kernel Learning}
Next we discuss the convergence rate of excess $\ell$-risk for kernel learning. We denote $\bdf=(f_1,\dots,f_{k-1})^T$ to be a function in a reproducing kernel Hilbert space (RKHS) $H$ with kernel function $K(\cdot, \cdot)$. Then by the RKHS theory, we can write $f_j(\bx)=\sum_{i=1}^nK(\bx_i, \bx)\alpha_{ij}+\alpha_{0j}$ for $j=1,\dots,k-1$. To develop the theory for the proposed methods, we still need one more assumption.

\begin{assumption}\label{asp:ker_bd}
Suppose $H$ is a separable RKHS equipped with kernel function $K(\cdot, \cdot)$. There exists a positive number $B$, such that $K(\bx, \bx')\leq B$ for any $\bx,\ \bx'\in\mX$.
\end{assumption}

Assumption \ref{asp:ker_bd} states that the RKHS is separable and the kernel function is bounded. This is true for many commonly used kernel functions. For example, for the Gaussian kernel, we may take $B=1$. 
We define the function space as
$$\mF(n, s)=\{\bdf=(f_1,\dots,f_{k-1})^T;f_j(\bx)=\sum_{i=1}^nK(\bx_i, \bx)\alpha_{ij}+\alpha_{0j}, j=1,\dots,k-1, J(\bdf)\leq s\},$$
where $J(\bdf)=\sum_{j=1}^{k-1}\balpha_j^TK\balpha_j+\sum_{j=1}^{k-1}\alpha_{0j}^2$ and $K$ is the gram matrix. Recall we have included intercepts in the penalty for simplicity. Let $\mF(\infty)=\lim_{n\to\infty}\cup_{0\leq s<\infty}\mF(n, s)$, and define
\begin{align*}
\bdf^{(\infty)}&=\argmin_{\bdf\in\mF(\infty)}\E\left[\frac{Y}{p(A|\bX)}\ell(\<{\bW_A, \bdf})\right],\\
\bdf^{(n, s)}&=\argmin_{\bdf\in\mF(n, s)}\E\left[\frac{Y}{p(A|\bX)}\ell(\<{\bW_A, \bdf})\right],\quad\text{and}\\
\hat\bdf&=\argmin_{\bdf\in\mF(n, s)}\frac{1}{n}\sum_{i=1}^n\frac{y_i}{p(a_i|\bx_i)}\ell(\<{\bW_{a_i}, \bdf(\bx_i)}),
\end{align*}
The following theorem gives the convergence rate of $e_\ell(\hat\bdf, \bdf^{(\infty)})$ when $s=s_n$ grows with $n$.

\begin{theorem}\label{thm:ker}
For RKHS learning, suppose Assumptions \ref{asp:reg}, \ref{asp:0measure}, \ref{asp:wt_bd}, and \ref{asp:ker_bd} hold. We have
$$e_\ell(\hat\bdf, \bdf^{(\infty)}) = O\left(\max(cB(s_n/n)^{1/2}\ln n, \delta_n)\right),$$
almost surely under $P$, where $\delta_n = e_\ell(\bdf^{(n, s_n)}, \bdf^{(\infty)})$.
\end{theorem}

Similar to the linear case, there is a trade-off between the approximation error $\delta_n$ and the estimation error $O(cB(s_n/n)^{1/2}\ln n)$ in Theorem \ref{thm:ker}, and the optimal tuning parameter $s_n$ is determined roughly when $cB(s_n/n)^{1/2}\ln n\sim\delta_n$.

Compared to Theorem \ref{thm:lin}, the excess $\ell$-risk for RKHS learning seems to yield a faster rate. However, this is not always truely the case due to Assumption \ref{asp:ker_bd}, which requires a bounded kernel function, and implies a restriction on the number of covariates $p$. For example, for linear kernel we have $K(\bx, \bx')=\bx^T\bx'\leq p$ under Assumption \ref{asp:x_bd}. For Assumption \ref{asp:ker_bd} to be true, we have to let $p=O(1)$. In this case both convergence rates are $e_\ell(\hat\bdf, \bdf^{(p)})=O((s_n/n)^{1/2}\ln n)$; that of the kernel learning is no faster than that of the linear learning. In general, to obtain a faster rate than that of the linear learning, we need a kernel function that does not increase in $p$, such as the Gaussian kernel.

Note that the approximation error $\delta_n$ converges to 0 as $n$ increases, and both the convergence rate of $\delta_n$ and that of the resulting $e_\ell(\hat \bdf, \bdf^*)$ depend on the choice of the kernel. To illustrate the magnitude of $\delta_n$ and its impact on the excess risk, consider a binary example where $X\sim\mathrm{Unif}(0, 1)$ and $f^*(x)=(1+x)^2$. With the polynomial kernel of degree 2 we have $f^{(\infty)}=f^*$ and $B=\max_{x,x'}(1+xx')^2=2$. Given a training set $\{x_1,\dots,x_n\}$, let $x_{(n)}$ be the largest order statistic and define $f_{(n)}(x)=\left(1+xx_{(n)}\right)^2$. It can be shown that for any $s_n\geq1$, $f_{(n)}\subseteq\mF(n, s_n)$ thus $\delta_n=e_\ell(f^{(n, s_n)}, f^{(\infty)})\leq e_\ell(f_{(n)}, f^{(\infty)})\leq c\E||f_{(n)}-f^{(\infty)}||_2$. Note that the difference between $f_{(n)}$ and $f^{(\infty)}$ is maximized at 1, so $\delta_n\leq c\E|f_{(n)}(1)-f^{(\infty)}(1)|=c\E\left(3-2x_{(n)}-x_{(n)}^2\right)$. Because the density function of $x_{(n)}$ is $nx^{n-1}\bone_{(0, 1)}$, we have $\delta_n\leq c\int_0^1(3-2x-x^2)nx^{n-1}dx=\frac{2c(2n+3)}{(n+1)(n+2)}$. Hence in this example, the order of $\delta_n$ is at most $O(n^{-1})$, thus $e_\ell(\hat f, f^*)=e_\ell(\hat f, f^{(\infty)})=O(n^{-1/2}\ln n)$.

\section{Simulation Studies}\label{sec:sim}
In this section, we study the numerical performance of the proposed methods.

\subsection{Comparing Set-valued Recommendations}
For two ITRs $d_1$ and $d_2$, we can compare them by evaluating the expected outcome defined in (\ref{E^d}). However, for two A-ITRs $\phi_1$ and $\phi_2$, it is difficult to quantify which one is better due to the fact that a measure for the overall benefit is not well defined when multiple treatments are recommended. Although in Section \ref{sec:tune} we have proposed the weighted expected outcome (\ref{wt_loss}) for evaluating two A-ITRs, the optimal A-ITR $\phi^+$ under this new criterion is still different from the desired near-optimal recommendation set $\phi^*$. So in the simulation studies, in addition to the empirical weighted outcome (\ref{empr_wt_loss}), we consider another means to compare different A-ITRs, using the expected outcome of the best and the worst treatments among the treatments that are recommended, averaged over a set of observations. We conduct such an evaluation for different types of recommendations separately to see how the A-ITR performs differently on them. Based on the size of the true optimal A-ITR $\phi^*$, we split the covariate space $\mX$ into three regions corresponding to three kinds of recommendations:
\begin{align*}
R_1&=\{\text{only one treatment is suggested}\}=\{\bx\in\mX;|\phi^*(\bx)|=1\},\\
R_2&=\{\text{more than one treatment but not all of them are suggested}\}\\
&=\{\bx\in\mX;1<|\phi^*(\bx)|<k\},\\
R_3&=\{\text{all treatments are suggested}\}=\{\bx\in\mX;|\phi^*(\bx)|=k\}.
\end{align*}
Note that $R_1$, $R_2$ and $R_3$ are disjoint and $\mX=R_1\cup R_2\cup R_3$. When $c=1$, $\phi^*$ is the optimal ITR $d^*$ and $\mX=R_1$. When $c>1$, we may have non-empty regions $R_2$ and $R_3$.

For two A-ITRs $\phi_1$ and $\phi_2$, we will compare them separately on $R_1$, $R_2$ and $R_3$. In each region, since multiple treatments may be suggested, we can compare the expected minimal outcome and the expected maximal outcome that they may lead to. Recall $Y^*(j)$ is the potential outcome by taking treatment $j$. Mathematically, we consider a performance interval,
$$\left(\E\left[\E\left(\min_{j\in\phi(\bX)}Y^*(j)|\bX\right)\right], \E\left[\E\left(\max_{j\in\phi(\bX)}Y^*(j)|\bX\right)\right]\right),$$
where the first quantity indicates the expected outcome if one can always use the best treatment within the recommended set $\phi(\bx)$ and the second quantity represents the worst situation, \ie, how bad it can be if one always chooses the worst treatment among the recommended options. Note that on $R_1$, the two quantities are the same under $\phi^*$ since only one treatment is recommended. As we increase $c$, we expect that this interval becomes wider on $R_2$ and $R_3$ since the diversity of the recommended options increases. From the definition of this interval, we claim that $\phi_1$ is better than $\phi_2$ if both the lower and the upper limits of this interval under $\phi_1$ are smaller than their counterparts under $\phi_2$, on each region.

\subsection{Results}
To compare the performance of our proposed methods, we consider three simulation examples. For each example, we consider two different dimensions with $\bX$ uniformly sampled from $\mX=[0, 1]^5$ and $\mX=[0, 1]^{10}$. For simplicity, we assume $A\perp\bX$ and $p(A|\bX)=1/k$, and let $Y=\mu_A(\bX)+\epsilon$ where $\epsilon\sim N(0, 1/2)$. In each case, we first generate a training sample with sample size 2000 to fit the model, then use a test set with sample size 1000 to evaluate the performance. We compare three methods, namely, the regression-based method, the two-step classification-based method with squared loss, and the one-step classification-based method with the bent hinge loss. For each method, we output both ITR and A-ITR with $c=1.2$. Finally, we repeat each simulation for 100 times and report the averages.

\begin{figure}[!htb]
\begin{center}
\includegraphics[width=5.5in]{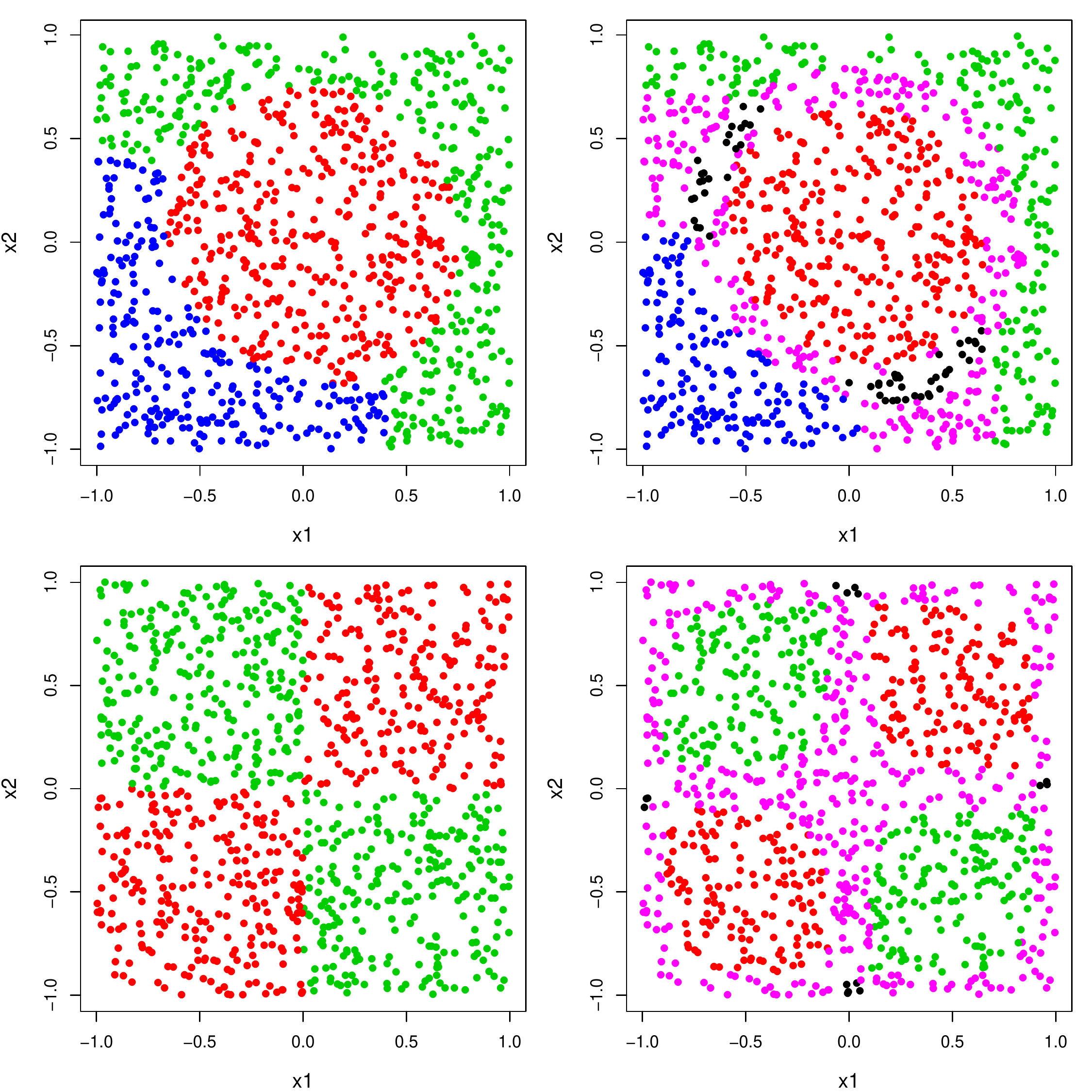}\vspace{-1.5em}
\end{center}
\caption{\small The true ITR and A-ITR for observations in Examples 1 and 2 with recommendation types shown in colors. The red, green and blue regions indicate that only one single treatment is suggested, which by definition is $R_1$, while the pink region is $R_2$ and the black region is $R_3$. The upper panel is Example 1, the lower panel Example 2. The left panel is ITR, and the right panel is A-ITR ($c=1.2$).}\vspace{-1em}\label{fig:sim}
\end{figure}

\textbf{Example 1:} This is an example with three treatments, where two conditional mean outcome functions are polynomial and the other is linear. Specifically, we have $\mu_1(\bX)=1+3X_1^2+3X_2^2$, $\mu_2(\bX)=3-0.5X_1^2+0.5X_2^2$, and $\mu_3(\bX)=3+X_1+X_2$. The upper panel in Figure \ref{fig:sim} shows the true boundaries for the three treatments. We report the results using polynomial kernel for both the two-step and one-step methods. The tuning parameter $\lambda$ is chosen from $5^{-6}$ to $5^2$.

\textbf{Example 2:} This is an example with four treatments, where all the conditional mean outcome functions are non-linear, $\mu_A(\bX)=2+\sign(A-2.5)\cos\left(0.5\pi(X_1+(-1)^AX_2)\right)$. Specially, treatments 2 and 4 are dominated by treatments 1 and 3 and the optimal ITR should only output either 1 or 3. However, in certain regions treatments 2 and 4 still produce fairly good outcome which can only be captured by A-ITR. The lower panel in Figure \ref{fig:sim} shows the true boundaries. For the two-step method, we report the results using Gaussian kernel. For the one-step method, we report the results with polynomial kernel. The tuning parameter $\lambda$ is chosen from $5^{-9}$ to $5^{-1}$.

\textbf{Example 3:} This is an example with three treatments and the true conditional mean outcome functions are determined by four covariates. Specifically, $\mu_1(\bX)=3+3X_1^2+3X_2^2-0.5\exp(0.5X_3^2+X_4)$, $\mu_2(\bX)=3-2X_1^2+\exp(X_3+X_4^2)$, and $\mu_3(\bX)=3-X_2^3-2X_3^2+0.5\left(\exp(X_1+X_4)-1\right)^2$. Similar to Example 1, we report the results using polynomial kernel for both the two-step and one-step methods. The tuning parameter $\lambda$ is chosen from $5^{-7}$ to $1$.

Table \ref{tbl:sim} collects the results of the three examples with $p=5$ and $10$. In Table \ref{tbl:sim}, the results of A-ITR are in the form of intervals while the results of ITR are single numbers. We also compute the empirical weighted outcome (``All" column in Table \ref{tbl:sim}) defined in (\ref{empr_wt_loss}) as an indicator for the overall performance for each method.

\begin{table}[!htb]
\footnotesize
\centering
\def\arraystretch{0.98}
\setlength\tabcolsep{2.9pt}
\begin{tabular}{cl|cccc|cccc}
\hline\hline
\multicolumn{2}{c}{\multirow{2}{*}{Example 1}}&\multicolumn{4}{c}{$p=5$}&\multicolumn{4}{c}{$p=10$}\\
\multicolumn{2}{c}{}&$R_1(70.05\%)$&$R_2(24.62\%)$&$R_3(5.33\%)$&All&$R_1(70.05\%)$&$R_2(24.62\%)$&$R_3(5.33\%)$&All\\
\hline
\multirow{2}{*}{Reg.}&ITR&2.47&2.52&2.61&2.49&2.47&2.53&2.62&2.50\\
&A-ITR&$(1.98, 3.09)$&$(2.42, 2.99)$&$(2.47, 2.96)$&2.37&$(1.99, 3.07)$&$(2.42, 2.99)$&$(2.48, 2.95)$&2.38\\
\hline
\multirow{2}{*}{2-step}&ITR&2.17&2.63&2.68&2.31&2.26&2.66&2.69&2.39\\
&A-ITR&$(2.03, 2.49)$&$(2.50, 2.83)$&$(2.57, 2.82)$&2.28&$(2.09, 2.52)$&$(2.54, 2.82)$&$(2.60, 2.78)$&2.32\\
\hline
\multirow{2}{*}{1-step}&ITR&2.07&2.59&2.67&2.23&2.21&2.65&2.68&2.34\\
&A-ITR&$(1.99, 2.23)$&$(2.51, 2.70)$&$(2.61, 2.74)$&\textbf{2.19}&$(2.08, 2.42)$&$(2.56, 2.76)$&$(2.62, 2.75)$&\textbf{2.30}\\
\hline
\multirow{2}{*}{Bayes}&ITR&1.92&2.44&2.56&2.08&&&&\\
&A-ITR&1.92&$(2.38, 2.74)$&$(2.46, 2.99)$&2.05&&&&\\
\hline\hline
\multicolumn{2}{c}{\multirow{2}{*}{Example 2}}&\multicolumn{4}{c}{$p=5$}&\multicolumn{4}{c}{$p=10$}\\
\multicolumn{2}{c}{}&$R_1(56.80\%)$&$R_2(41.84\%)$&$R_3(1.36\%)$&All&$R_1(56.80\%)$&$R_2(41.84\%)$&$R_3(1.36\%)$&All\\
\hline
\multirow{2}{*}{Reg.}&ITR&1.56&1.62&1.92&1.60&1.56&1.62&1.93&1.60\\
&A-ITR&$(1.13, 1.99)$&$(1.21, 2.04)$&$(1.78, 2.09)$&1.45&$(1.14, 1.99)$&$(1.21, 2.03)$&$(1.80, 2.09)$&1.46\\
\hline
\multirow{2}{*}{2-step}&ITR&1.24&1.30&1.93&1.28&1.36&1.36&1.93&1.37\\
&A-ITR&$(1.15, 1.48)$&$(1.22, 1.46)$&$(1.83, 2.06)$&1.26&$(1.23, 1.57)$&$(1.27, 1.54)$&$(1.87, 2.03)$&1.35\\
\hline
\multirow{2}{*}{1-step}&ITR&1.24&1.30&1.93&1.28&1.32&1.34&1.92&1.34\\
&A-ITR&$(1.16, 1.41)$&$(1.23, 1.39)$&$(1.85, 2.01)$&\textbf{1.25}&$(1.20, 1.50)$&$(1.26, 1.45)$&$(1.87, 1.99)$&\textbf{1.31}\\
\hline
\multirow{2}{*}{Bayes}&ITR&1.13&1.25&1.87&1.19&&&&\\
&A-ITR&1.13&$(1.15, 1.46)$&$(1.71, 2.30)$&1.16&&&&\\
\hline\hline
\multicolumn{2}{c}{\multirow{2}{*}{Example 3}}&\multicolumn{4}{c}{$p=5$}&\multicolumn{4}{c}{$p=10$}\\
\multicolumn{2}{c}{}&$R_1(58.79\%)$&$R_2(29.93\%)$&$R_3(11.28\%)$&All&$R_1(58.79\%)$&$R_2(29.93\%)$&$R_3(11.28\%)$&All\\
\hline
\multirow{2}{*}{Reg.}&ITR&3.35&3.57&3.85&3.47&3.35&3.59&3.86&3.49\\
&A-ITR&$(3.01, 3.85)$&$(3.41, 3.82)$&$(3.72, 4.01)$&3.35&$(3.01, 3.87)$&$(3.43, 3.83)$&$(3.72, 4.01)$&3.37\\
\hline
\multirow{2}{*}{2-step}&ITR&3.09&3.53&3.82&3.30&3.23&3.59&3.84&3.41\\
&A-ITR&$(2.89, 3.51)$&$(3.37, 3.80)$&$(3.68, 4.01)$&3.21&$(2.99, 3.61)$&$(3.44, 3.81)$&$(3.74, 3.96)$&\textbf{3.32}\\
\hline
\multirow{2}{*}{1-step}&ITR&3.05&3.50&3.82&3.26&3.23&3.59&3.85&3.41\\
&A-ITR&$(2.87, 3.43)$&$(3.37, 3.66)$&$(3.73, 3.92)$&\textbf{3.19}&$(2.96, 3.74)$&$(3.43, 3.80)$&$(3.75, 3.95)$&3.33\\
\hline
\multirow{2}{*}{Bayes}&ITR&2.72&3.28&3.65&2.99&&&&\\
&A-ITR&2.72&$(3.23, 3.66)$&$(3.58, 4.19)$&2.92&&&&\\
\hline\hline
\end{tabular}
\caption{\small Results of the simulation studies. In each region, the expected outcome for ITR and the outcome interval for A-ITR ($c=1.2$) are reported. The empirical weighted outcome defined in (\ref{empr_wt_loss}) is shown in the ``All" column. Each number is averaged over 100 replications. In each case, the best performing method is marked in bold.}\vspace{-1em}\label{tbl:sim}
\end{table}

We note that the performance intervals for A-ITR always cover the expected outcomes of the single-valued ITR. This implies that by applying our proposed A-ITR framework, patients will potentially get a much better outcome as long as they are willing to consider other equally effective options identified by the A-ITR. Even if the patient does not choose the best option within the recommended set, the worst case is not too bad and the ratio of its outcome to that of the best option is about $c$ if the A-ITR is accurately estimated. 

We compare different methods by inspecting the length and location of the A-ITR performance interval. Recall that the A-ITR with the shortest interval, the smallest lower limit, and the smallest upper limit on each region is the best A-ITR. However, since $R_3$ is the region where all treatments are near the optimality, different recommendations are expected to perform similarly. Hence we focus on regions $R_1$ and $R_2$ for the purpose of comparison.


From Table \ref{tbl:sim}, we note that the regression-based A-ITR, though has the smallest lower limit in some cases, always yields the longest interval, suggesting that the treatment could either go really well or really badly. This implies that the regression-based A-ITR method tends to include ineffective treatments into the near-optimal set. Part of the reason may be that the regression-based method has not accurately estimated each of the three or four potential outcome functions.

For the classification-based A-ITRs, the lower limits are roughly the same between the one-step method and the two-step method; however, the one-step method has shorter interval in most cases. This means that the one-step method is better at excluding ineffective treatment options from the recommendation than the two-step method. In addition, the one-step method also has the smallest expected weighted outcome (shown in the ``All" column).

\section{Real Data Analysis}\label{sec:real}
In this section, we apply our proposed A-ITR framework to a Type 2 diabetes mellitus (T2DM) observational study. The data set contains 1139 patients. Every patient was assigned with one out of four diabetes treatments, which are GLP-1 receptor agonists alone, long-acting insulin alone, intermediate-acting insulin alone, and insulin regimens including a short-acting insulin. The end point is the change of hemoglobin A1c level before and after the treatment, which is denoted by $\Delta HbA1c$. In practice, if the treatment works, this value is usually negative (meaning that the hemoglobin A1c level decreases). The smaller $\Delta HbA1c$ is, the more effective the treatment is.

We first preprocess the original data set. Among the 19 covariates, we exclude those with large proportion of missing values and with extremely imbalanced categories. We then impute the rest of them using the predictive mean matching method \citep{van2018flexible}. There are 10 covariates left after the preprocessing: gender, diabetic retinopathy, diabetic neuropathy, age, weight, body mass index (BMI), baseline hemoglobin A1c level, baseline high-density lipoprotein cholesterol (HDL), baseline low-density lipoprotein cholesterol (LDL), and heart disease.

For the outcome variable $\Delta HbA1c$, we can reduce its variability by subtracting an estimate of its conditional mean $\E(\Delta HbA1c|\bx)$ to make the estimation of $\hat\bdf$ more robust \citep{liu2016robust, zhou2017residual}. Here we use the ordinary least square regression to estimate $\E(\Delta HbA1c|\bx)$. Denote the estimated mean function fitted by regression as $\hat m(\bx)$, we then observe that $\Delta HbA1c-\hat m(\bx)$ can be positive or negative. We perform an exponential transformation to make it positive, which also justifies the use of ratio $\mu_j/\mu_{(1)}$ to determine the near-optimal recommendation set. Specifically, we let $Y=\exp\left((\Delta HbA1c-\hat m(\bx))/5\right)$. If we further assume conditional normality for $\Delta HbA1c$ given $\bX$ and treatment $j$, with mean $\nu_j(\bx)\equiv\E(\Delta HbA1c|\bX=\bx, A=j)$ and equal variance across treatments, then $Y|(\bx,j)$ follows a log-normal distribution with mean proportional to $\exp(\nu_j(\bx)/5)$. Then the optimal A-ITR is,
\begin{align*}
\qquad\phi^*(\bx)&=\left\{j;\frac{\mu_j(\bx)}{\mu_{(1)}(\bx)}=\frac{\E\left(Y|\bX=\bx, A=j\right)}{\min_i\E\left(Y|\bX=\bx, A=i\right)}=\frac{\exp(\nu_j(\bx)/5)}{\exp(\nu_{(1)}(\bx)/5)}\leq c\right\}\\
&=\{j;\nu_j(\bx)-\nu_{(1)}(\bx)\leq 5\log c\}.
\end{align*}

In this study, we choose the near-optimal parameter $c=1.2$, so that $5\log c\approx0.9$. This implies that the near-optimal recommendation set is constructed by including all treatments with conditional means $\Delta HbA1c$ within 0.9 of the optimal treatment.

We compare performance of the regression-based method, the two-step method, and the one-step method. For both classification-based methods, we estimate the propensity score $p(A|\bX)$ using logistic regression. Each method leads to a single-valued ITR and a set-valued A-ITR, and we compare the different recommendations using the 5-fold cross-validated empirical weighted outcome defined in (\ref{empr_wt_loss}), shown in Table \ref{tbl:real}.

\begin{table}[!htbp]
\footnotesize
\centering
\def\arraystretch{0.98}
\begin{tabular}{ccc}
\hline\hline
&ITR&A-ITR\\
\hline
\multirow{2}{*}{Regression}&1.071&0.988\\
&(0.010)&(0.006)\\
\hline
\multirow{2}{*}{Two-step: Linear}&0.995&0.975\\
&(0.007)&(0.006)\\
\hline
\multirow{2}{*}{Two-step: Gaussian}&0.959&0.947\\
&(0.006)&(0.005)\\
\hline
\multirow{2}{*}{One-step: Linear}&1.150&1.033\\
&(0.008)&(0.006)\\
\hline
\multirow{2}{*}{One-step: Gaussian}&\textbf{0.939}&\textbf{0.935}\\
&(0.006)&(0.007)\\
\hline\hline
\end{tabular}
\caption{\small The mean 5-fold cross validated weighted outcome and its standard error (in the parenthesis) over 100 replications for T2DM data. The method that yields the best result is marked in bold.}\vspace{-1em}\label{tbl:real}
\end{table}

\begin{figure}[!htb]
\begin{center}
\includegraphics[width=\linewidth]{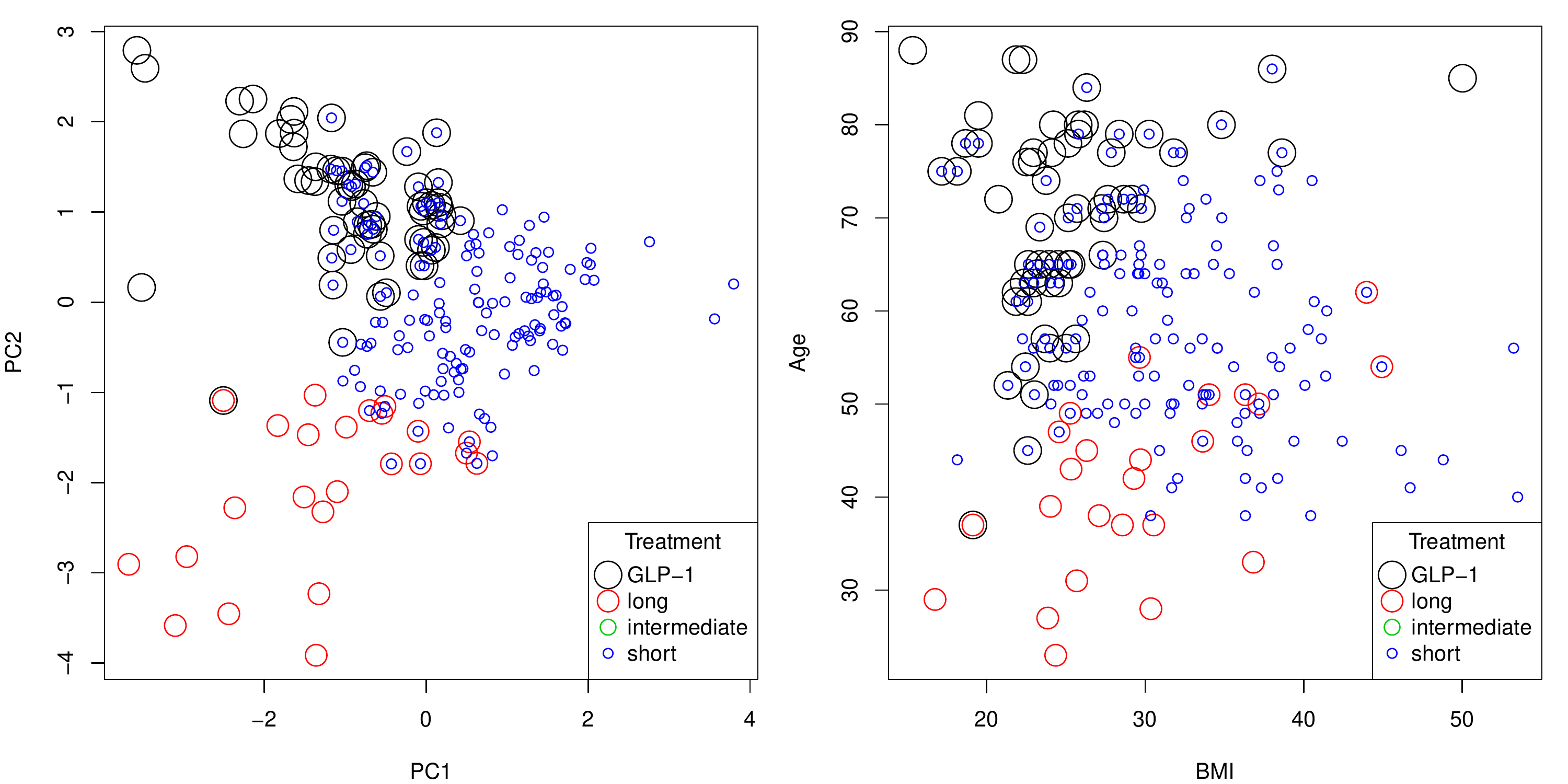}\vspace{-1.5em}
\end{center}
\caption{\small Predicted treatment(s) for patients with recommendations given by different colors. The data set is projected on the first two principle components (left panel), and two particular covariates, age and BMI (right panel). Concentric circles indicate multiple treatments recommended to the same patient.}\vspace{-1em}\label{fig:real}
\end{figure}

From Table \ref{tbl:real}, we observe that the one-step method with Gaussian kernel has the best weighted outcome. To illustrate the resultant A-ITR, we split the data into training set (70\%) and test set (30\%). We fit the training set using one-step method with Gaussian kernel and then construct the recommendation set for patients in the test set. In our analysis, no patient is recommended to take the intermediate-acting insulin and the majority of patients are recommended to choose between the short-acting insulin and GLP-1. Specifically, 55\% of patients are recommended the short-acting insulin only, 8\% are recommended GLP-1 only, and 24\% are recommended to take either one of the two. For the rest 13\% of patients, they are all recommended to take the long-acting insulin, including 1\% who are suggested to take either long-acting insulin or GLP-1, 5\% who are suggested to take either long-acting insulin or short-acting insulin, and 7\% whose only option is long-acting insulin. We visualize the predicted treatments in Figure \ref{fig:real}.

From Figure \ref{fig:real}, we can see that age and BMI are two useful biomarkers in constructing the near-optimal recommendation set. In fact, by comparing the left panel and the right panel of Figure \ref{fig:real}, we observe that BMI behaves like the first principle component (PC1) while age behaves like the second principle component (PC2). Figure \ref{fig:real} suggests that for patients without obesity (BMI less than 30), younger patients should take the long-acting insulin while older patients should take GLP-1. The short-acting insulin, on the other hand, serves as an ``universal" treatment that many patients can take as an alternative, and is especially effective for overweighted patients.

\section{Conclusion}\label{sec:con}
In this work, we propose a new individualized treatment recommendation framework, named A-ITR, that has the capacity to recommend to patients near-optimal treatment options in terms of their clinical outcomes. By adopting the A-ITR, patients have the opportunity to choose the treatment options tailored for their different financial situations, personal preference and life style choices. To estimate the optimal A-ITR, we proposed two classification-based methods based on the OWL framework. We also provide a new evaluation criterion suitable for A-ITRs, namely the weighted expected outcome, defined in (\ref{wt_loss}). The simulation study shows the usefulness of this new criterion in parameter tuning and model selection.



There are several possible directions for future works. Firstly, the current A-ITR estimation is based on OWL framework, which may be sensitive to the estimated propensity score. In this case, one may consider applying the doubly-robust OWL framework \citep{zhao2019efficient, huang2019multicategory} to improve the efficiency of the estimated A-ITR. Secondly, we can consider A-ITRs with an additional competing outcome as a secondary end point \citep{laber2014set}, or A-ITR with additional safety end points formulated as constraints \citep{wang2018learning}. Thirdly, we can consider other learning algorithms to estimate $\hat\bdf$ within the A-ITR framework. Finally, we can consider nontrivial extensions to other types of outcome such as survival outcome \citep{zhao2014doubly, qi2019multi} or dichotomous outcome \citep{qi2019multi, klausch2018estimating}.

\bibliographystyle{asa}
\bibliography{reference}

\end{document}